\documentclass[11pt, a4paper, logo, copyright, gdm]{google}
\usepackage[authoryear, sort&compress, round]{natbib}
\usepackage{times}

\usepackage{amsmath,amsfonts,bm}

\def\eqref#1{equation~\ref{#1}}

\def\1{\bm{1}}

\DeclareMathAlphabet{\mathsfit}{\encodingdefault}{\sfdefault}{m}{sl}
\SetMathAlphabet{\mathsfit}{bold}{\encodingdefault}{\sfdefault}{bx}{n}

\usepackage{hyperref}
\usepackage{url}
\usepackage{colortbl}
\usepackage{array}
\usepackage{makecell}
\usepackage{graphicx} %
\usepackage{amsthm}
\usepackage{amsmath}
\usepackage{subcaption}
\usepackage{booktabs}
\usepackage{multirow}   %
\usepackage{wrapfig}
\usepackage{amssymb}
\usepackage{tikz}
\usetikzlibrary{decorations.pathreplacing}

\usepackage{algorithm}
\usepackage{algpseudocode}

\uselogo{}
\title{RES: Accelerating Diffusion Language Models with Rolling Entropy Sampling}

\author[1,*]{Patrick Pynadath}
\author[2]{Nicolas Perez-Nieves}
\author[2]{Bobak Shahriari}
\author[2]{Jean Tarbouriech}
\author[1]{Ruqi Zhang}
\author[*]{Ivan Lobov}
\affil[1]{Purdue University}
\affil[2]{Google DeepMind}
\affil[*]{Work done while at Google DeepMind.}

\usepackage[section]{placeins}
\begin{document}

\newcommand{\algname}{RES}
\newcommand{\algabbrev}{RES}

\begin{abstract}
A diffusion language sampler makes two choices for every position in its denoising window: where the next token comes from, and when the position is committed to the KV cache.
Using the entropy-bound sampler in DiffusionGemma, we study two extremes of the second choice.
Committing each token as soon as it is accepted removes the chance to revise it, which hurts when the entropy bound is relaxed to trade accuracy for speed.
When accepted tokens are frozen, macro accuracy falls from 73\% to 20\% as the bound is relaxed, while the unfrozen sampler stays near 70\%.
Committing a whole block at once wastes computation, since tokens converge in a largely left-to-right order and more than half of the window holds already-final tokens by twenty denoising steps.
These findings motivate the Rolling Entropy Sampler (RES), which commits a contiguous accepted prefix once it reaches a minimum length and keeps positions farther ahead open to revision.
Unlike earlier rolling samplers, RES runs on uniform diffusion, where accepted tokens can still change, and it needs no training.
Across seven benchmarks, RES increases tokens per forward pass from 19.3 to 25.5, while macro accuracy goes from 80.7\% to 78.8\%, and it extends the speed-quality Pareto frontier beyond what tuning the entropy bound alone reaches.
\end{abstract}

\maketitle

\section{Introduction}
\label{sec:intro}
Autoregressive language models train on all tokens in parallel, yet must generate them one at a time.
Diffusion language models remove this limit by denoising many tokens in each forward pass \citep{nie2026large, ye2025dream7bdiffusionlarge, team2026diffusiongemma}.
Samplers are a natural place to improve these models: in continuous diffusion, many advances have come from better samplers \citep{song2020denoising}, and in language diffusion, the sampling order strongly affects task performance \citep{kim2025train, ben2026accelerated, kim2026klass}.

Any diffusion language sampler makes two choices for each position in its denoising window.
The first is \emph{where} the next token comes from: the model's prediction, the token already at that position, or fresh noise.
This defines when a position within a denoising window has converged.
The second is \emph{when} the position is committed to the KV cache, after which it can no longer change.
These design choices give rise to two potential extremes: on one end, it is possible to commit tokens on a \textit{per position} basis, or as soon as a position has converged.
It is also possible to commit tokens \textit{at a block level}, or after the entire denoising window has converged.

In this work, we use the Entropy Bound sampler used in DiffusionGemma \citep{team2026diffusiongemma} to study these extremes and uncover the underlying tradeoffs.
Committing each token as soon as it converges removes the chance to revise it later.
Revision matters most when the entropy bound is relaxed to trade accuracy for speed.
When accepted tokens are frozen, macro accuracy falls from 73\% to 20\% as the entropy bound is relaxed, while the unfrozen sampler stays near 70\%.
Committing on a block level results in many redundant computations, as tokens converge in a largely left-to-right order.
Thus a converged prefix sits idle in the window, and by twenty denoising steps more than half of the slots, on average, are spent on tokens that are already final.

Since per-position commitment gives up revision and block-level commitment wastes computation, we introduce the \textbf{R}olling \textbf{E}ntropy \textbf{S}ampler (RES).
RES commits a contiguous prefix of accepted positions once that prefix holds at least $\ell_{\min}$ positions, and it keeps positions farther ahead open to revision.
Earlier rolling samplers work with masked tokens, which are frozen once accepted.
DSB \citep{luo2026dsb} rolls the window for masked diffusion, and Set Block Decoding \citep{gat2025set} fine-tunes an autoregressive model to decode masked tokens ahead.
Concurrent work on Clock Diffusion \citep{schiff2026clock} instead trains a Gaussian diffusion model with a noise schedule that depends on position.
RES differs from both in two ways.
First, it runs on uniform diffusion, where every position always holds a token that can still change. Section~\ref{sec:too-early} shows that keeping this ability matters.
Second, RES changes only inference, so it needs no additional training.

In summary, our contributions are as follows:
\begin{itemize}
    \item Separating accepting a token from committing it (Section~\ref{sec:motivation}), we show that DiffusionGemma revises its own accepted tokens without renoising, and that freezing them lowers macro accuracy from 73\% to 20\% at loose entropy bounds (Section~\ref{sec:too-early}).
    \item We show that the EB sampler converges left to right, leaving over half of the window idle late in each block (Section~\ref{sec:too-late}).
    \item We introduce RES, a rolling sampler that needs no training (Section~\ref{sec:method}). Across seven benchmarks it increases tokens per forward pass from 19.3 to 25.5, while macro accuracy goes from 80.7\% to 78.8\%. It also extends the speed-quality frontier beyond what relaxing the entropy bound alone reaches (Section~\ref{sec:experiments}).
\end{itemize}

\section{Preliminaries}
\label{sec:prelims}
We study uniform diffusion with a denoising window and KV caching. We review its corruption process and entropy-bound acceptance, and include masked diffusion as background for comparing sampler rules.
\subsection{Discrete Diffusion}
Here we cover the relevant background for discrete diffusion.
When referring to a sequence, we will use upper-case $X$ whereas individual tokens will be represented by $x$, written as one-hot vectors over a vocabulary of size $K$.
We write $x_0$ for a clean token and $x_t$ for its noisy version at time $t \in [0, 1]$, and we take $s < t$, so that $x_s$ is a less noisy version of $x_t$.
The keep rate $\alpha_t \in [0, 1]$ decreases from $\alpha_0 = 1$ to $\alpha_1 = 0$, and $\alpha_{t|s} = \alpha_t / \alpha_s$.
Finally, $m$ denotes the one-hot mask token, $\mathbf{1}_K$ the all-ones vector, and $\langle \cdot, \cdot \rangle$ the inner product.

\paragraph{Uniform Diffusion}
Uniform diffusion replaces clean tokens with tokens drawn uniformly at random from the vocabulary \citep{schiff2025simple, sahoo2025diffusion}.
The reverse bridge distribution is defined as follows:
\begin{align}
\label{eq:bridge}
    q_{s|t}(x_s | x_t, x_0) = \text{Cat} \left( \frac{x_t \cdot (K\alpha_t \langle x_t, x_0 \rangle + \alpha_{t|s} - \alpha_t) + x_0 \cdot ( \alpha_s - \alpha_t) + D_{s,t} \mathbf{1}_K / K}{K \alpha_t \langle x_t, x_0 \rangle + 1 - \alpha_t} \right),
\end{align}
where $D_{s,t} = (1 - \alpha_{t|s})(1 - \alpha_s)$.

To use this for generation, it is common to either substitute a neural network for $x_0$ as a plug-in parameterization, or to learn the denoiser $P(X_0 | X_t)$. As shown in \citet{gourevitch2026uniform}, it is possible to translate between these parameterizations. Thus we assume the denoiser parameterization.

To sample from the bridge, we first sample $X_0 \sim P_{\theta}(\cdot | X_t)$, $U \sim \text{Cat}(\frac{1}{K})$.
For each position, we sample from $\text{Cat}(p_{accept} \cdot x_0 + p_{stay} \cdot x_t + p_{noise} \cdot u)$ with probabilities defined as follows:
\begin{align}
\label{eq:true-rev-coords}
    p_{accept} &= \frac{\alpha_s - \alpha_t}{Z(x_t, x_0)} \\
    p_{stay} &= \frac{K\alpha_t \langle x_t, x_0 \rangle + \alpha_{t|s} - \alpha_t}{Z(x_t, x_0)} \\
    p_{noise} &= \frac{D_{s,t}}{Z(x_t, x_0)} \\
    Z(x_t, x_0) &= K \alpha_{t} \langle x_t, x_0 \rangle + 1 - \alpha_t
\end{align}

DiffusionGemma also uses self conditioning \citep{chen2022analog}. At each step, the model receives its own previous prediction as an extra input, so every position carries information from earlier steps.

\paragraph{Masked Diffusion}
Masked diffusion incrementally corrupts a sequence by replacing tokens in sequence $X$ with a predefined ``mask'' token, which serves as the absorbing state for the process \citep{campbell2022continuous, sahoo2024simple, shi2024simplified}.
Given a keep rate $\alpha_t$, the corruption process is defined as follows:
\[
q_{t|0}(\cdot | x_0) \approx \text{Cat} \left( \alpha_t \cdot x_0 + (1 - \alpha_t) \cdot m\right).
\]
Masked diffusion has seen success when scaling up diffusion language models \citep{nie2026large, ye2025dream7bdiffusionlarge}.
In monotone masked-diffusion samplers, an unmasked position remains fixed; methods that enable subsequent revisions have also been studied \citep{kim2025fine}. This freezing behavior is distinct from moving a position into a KV cache. We discuss masked diffusion for context; all experiments in this paper use the same uniform-diffusion model.

\paragraph{Block Diffusion}
Block diffusion is a technique that interpolates between autoregressive order and pure diffusion. Given a maximum context length $C$, block diffusion splits this into evenly sized contiguous blocks $b$ that are denoised in a left-to-right order \citep{arriola2025blockdiffusioninterpolatingautoregressive, deschenaux2026blockgen}.
This technique is useful as it allows for some form of KV caching whereas pure diffusion requires recomputing attention across all positions every step.

\subsection{Adaptive Samplers}
Adaptive acceptance rules use model predictions to choose which positions to update. Prior work on masked diffusion demonstrates the importance of generation order \citep{kim2025train}.
Towards this end, training-free adaptive sampling algorithms are commonly used as they require no architectural modifications \citep{nie2026large, ye2025dream7bdiffusionlarge}.
In general, such algorithms typically rely on the model's logits to determine the ordering, although some also leverage the internal attention patterns \citep{kim2026klass, wu2026fast}.

\paragraph{Entropy Bound Sampler}
The entropy-bound acceptance rule was introduced for masked-diffusion sampling by \citet{ben2026accelerated} and is used in the uniform-diffusion sampler of DiffusionGemma \citep{team2026diffusiongemma}. Using this criterion does not impose a masked corruption process or require accepted positions to be frozen.
As our study primarily investigates DiffusionGemma, we briefly review the EB sampler.
The EB sampler sorts tokens by the predictive entropy in ascending order and accepts the largest subset $U$ such that:
\begin{align}
    \label{eq:eb-bound}
    \sum_{i \in U} H(P_{\theta}^i(\cdot | X_t)) - \max_{i \in U}  H(P_{\theta}^i(\cdot | X_t)) \le \gamma.
\end{align}
Here, $\gamma \ge 0$ is a tunable hyperparameter, the entropy bound. As discussed in \citet{ben2026accelerated}, this sampler approximately bounds joint dependence error when the model has low prediction error.

\section{Where and When: Two Choices for Every Position}
\label{sec:motivation}
In this section, we write existing samplers in terms of a routing rule and a commitment rule (Algorithm~\ref{alg:gen-adapt}, Table~\ref{tab:samplers}).
We assume a diffusion model that learns the posterior $P(X_0 | X_t)$, and takes both a denoising window and a KV cache as input.
The denoising window represents tokens that are actively being denoised, and the KV cache corresponds to tokens that have been completely denoised.

At every time-step, the sampler must answer the following two questions:
\begin{enumerate}
    \item \textit{Where does the token come from?} Is it the predicted token $\hat{X}_i \sim P_{\theta}(\cdot | X_t)_i$, the previous value $X_t^i$, or a fresh noisy token drawn from $\pi$?
    \item \textit{When does the position get appended to the KV cache?} Do we move the position to the KV cache, or keep it in the denoising window?
\end{enumerate}

To answer the first question, a sampler defines a routing rule $F_{route}$ that specifies a point $(p_{accept},\allowbreak p_{stay},\allowbreak p_{noise})$, which we refer to as the accept-stay-noise simplex.
A position is \emph{accepted} when the routing rule takes the model's prediction for the current step.

The second question is answered by a commitment rule $F_{commit}$ that moves positions from the denoising window to the KV cache, after which their tokens can no longer change.
Tying commitment to each position's acceptance entangles these choices; a block or prefix rule instead waits for a set of positions to be accepted.
The subscripts in $p_{accept}$ and $F_{commit}$ denote acceptance and cache commitment, respectively.

\begin{algorithm}[tb]
\caption{Generic Diffusion Sampler}
\label{alg:gen-adapt}
\begin{algorithmic}[1]
\Require Posterior $P_\theta$, encoder prompt $x_{\mathrm{prompt}}$, window size $w$, noise prior $\pi$, routing rule $F_{\mathrm{route}}(*) \to (\Delta_{3})^w = (p_{accept}, p_{stay}, p_{noise})$, commit rule $F_{\mathrm{commit}}(*) \to \{0, 1\}^w$
\State $X_{\mathrm{done}} \gets x_{\mathrm{prompt}}$
\State $X \sim \pi(V^w)$
\While{\textsc{Eos} not committed}
    \State $\hat{X} \sim P_{\theta}(\cdot \mid X, X_{\mathrm{done}})$
    \State $U \sim \pi(V^w)$
    \State $[p_{\mathrm{accept}}^i, p_{\mathrm{stay}}^i, p_{\mathrm{noise}}^i]_1^w \gets F_{\mathrm{route}}(X, \hat{X}, P_{\theta}, *)$
    \State $X_i \sim \text{Cat} \left( p_{\mathrm{accept}}^i \cdot \hat{X}_i + p_{\mathrm{stay}}^i \cdot X_i + p_{\mathrm{noise}}^i \cdot U_i \right)$
    \State $[c_1, \dots, c_w] \gets F_{\mathrm{commit}}(X, \hat{X}, P_{\theta}, *)$ \Comment{Binary readiness for each position}
    \State $X_{\mathrm{done}} \gets X_{\mathrm{done}} \cup \{X_{i} : c_i = 1\}$ \Comment{Commit contiguous prefix}
    \State $X \gets \{X_{i}: c_i = 0 \} \cup X_{\mathrm{new}}$, $X_{\mathrm{new}} \sim \pi(V^{n})$, $n = \sum_i c_i$ \Comment{Roll window and fill tail with fresh noise}
\EndWhile
\State \Return $X_{\mathrm{done}}$
\end{algorithmic}
\end{algorithm}

\newcommand{\ind}[1]{\mathbb{1}\!\left[#1\right]}
\newcommand{\Aeb}{A_{\gamma}}                 %
\newcommand{\Aconf}{A_{\tau}}                 %
\newcommand{\Aever}{A^{<t}}                   %
\newcommand{\pref}[1]{\ell(#1)}               %
\newcommand{\post}{P_{\theta}}
\newcommand{\Hbar}{\bar{H}}
\newcommand{\ours}[1]{\cellcolor{gray!20}#1}
\newcommand{\samp}[2]{\makecell[l]{#1\\[-2pt]{\scriptsize #2}}}

\begin{table}[tb]
\centering
\small
\setlength{\tabcolsep}{4pt}
\renewcommand{\arraystretch}{1.1}
\resizebox{\linewidth}{!}{%
\begin{tabular}{@{}l c l l l l@{}}
\toprule
Sampler & $q$ & $\hat X \sim$ & $p^{i}_{\text{accept}}$ & $p^{i}_{\text{stay}}$ & $c_i$ \\
\midrule
\multicolumn{6}{@{}l}{\textit{Commit never / at block end}} \\
Ancestral (uniform) & unif & $\post(\cdot\mid X_{\text{done}})$ & Eq.~(\ref{eq:true-rev-coords}) & Eq.~(\ref{eq:true-rev-coords}) & $\ind{t=0}$ \\
Ancestral (masked) & mask & $\post(\cdot\mid X_{\text{done}})$ & $\ind{X_i = m}\,\tfrac{\alpha_s-\alpha_t}{1-\alpha_t}$ & $1-p^{i}_{\text{accept}}$ & $\ind{t=0}$ \\
\samp{Confidence}{\citep{nie2026large,wu2026fast}} & mask & $\post$ & $\ind{i\in\Aconf \wedge X_i=m}$ & $1-p^{i}_{\text{accept}}$ & $\ind{m\notin X}$ \\
\samp{EB}{\citep{ben2026accelerated}} & mask & $\post$ & $\ind{i\in\Aeb \wedge X_i=m}$ & $1-p^{i}_{\text{accept}}$ & $\ind{m\notin X}$ \\
\samp{Set Block Decoding}{\citep{gat2025set}} & AR & $\post(\cdot\mid X_{\text{done}}, m,\dots,m)$ & $\ind{i\in\Aeb \wedge X_i=m}$ & $1-p^{i}_{\text{accept}}$ & $\ind{m\notin X}$ \\
\samp{DiffusionGemma block EB}{\citep{team2026diffusiongemma}} & unif & $\post$ & $\ind{i\in\Aeb}$ & $0$ & $\ind{\text{stable}_s \wedge \Hbar<\tau_H}$ \\
Unified EB (ours) & unif & $\post$ & $\ind{i\in\Aeb}$ & $0$ \,/\, $1-p^{i}_{\text{accept}}$ (ablated) & \ours{$\ind{\pref{\Aeb}=w}$} \\
Frozen EB (ablation) & unif & $\post$ & $\ind{i\in\Aeb\setminus\Aever}$ & $\ind{i\in\Aever}$ & $\ind{\pref{\Aeb}=w}$ \\
\midrule
\multicolumn{6}{@{}l}{\textit{Commit incrementally}} \\
\samp{Jacobi / Lookahead}{\citep{santilli2023accelerating,fu2024break}} & AR & $\arg\max \post(\cdot\mid X_{\text{done}},X_{<i})$ & $1$ & $0$ & $\ind{\hat X_j = X_j\ \forall j\le i}$ \\
\samp{Adaptive Parallel Decoding}{\citep{israel2026accelerating}} & mask & $\post$ & $\ind{i\le\pref{A_{\text{mix}}}}$ & $1-p^{i}_{\text{accept}}$ & $\ind{i\le\pref{A_{\text{mix}}}}$ \\
\samp{DSB}{\citep{luo2026dsb}} & mask & $\post$ & $\ind{i\in\Aconf \wedge X_i=m}$ & $1-p^{i}_{\text{accept}}$ & $\ind{i\le\pref{\Aconf\cup\{j:X_j\neq m\}}}$ \\
RES (ours) & unif & $\post$ & $\ind{i\in\Aeb}$ & $0$ \,/\, $1-p^{i}_{\text{accept}}$ & \ours{$\ind{i\le\pref{\Aeb}}\,\ind{\pref{\Aeb}\ge \ell_{\min}}$} \\
\bottomrule
\end{tabular}%
}
\caption{%
Samplers as instances of Algorithm~\ref{alg:gen-adapt}: proposal $\hat X$, accept/stay probabilities ($p_{\text{noise}} = 1 - p_{\text{accept}} - p_{\text{stay}}$), and commit rule $c_i$.
$\Aeb$ is the entropy-bound set (Eq.~(\ref{eq:eb-bound})), $\Aconf$ the confidence set, $\pref{A}$ the longest accepted prefix, $\Aever$ the positions accepted at earlier steps.
Shaded cells highlight the commitment rules introduced here; RES with minimum shift $\ell_{\min}=w$ recovers Unified EB when the remaining sampler settings, including the temperature schedule, are matched.
Freezing is encoded by a rule that forces future stay probabilities to one, whereas $c_i$ denotes cache commitment. Frozen EB simulates per-position commitment by preventing later revision; RES waits for a contiguous accepted prefix and leaves accepted positions farther ahead revisable.
}
\label{tab:samplers}
\end{table}

\paragraph{The Accept-Stay-Noise Simplex}
The reverse bridge in equation~\ref{eq:true-rev-coords} gives the probabilities of these three routing choices, with $\pi$ as the prior distribution of the process.
We borrow this routing view from RDM \citep{zheng2023reparameterized}, which introduces auxiliary routing variables and makes the available moves depend on whether the current token is already clean: a clean token either stays or is renoised, and a noisy token is either denoised or renoised.
We introduce no auxiliary variables and place all three moves on a single simplex: the expectation of the clean-token indicator is the posterior's mass on the current token, and the bridge already exposes the three moves without it.

This formulation also matches the design of HackableDiffusion, an open-source framework for diffusion models \citep{hackable_diffusion2026github}.

\paragraph{Adaptive Samplers as Routing Rules}
The routing view was introduced to make this choice explicit: \citet{zheng2023reparameterized} observe that the standard sampler routes every token with the same probability, and that writing the router out allows more flexible decoding rules.
Adaptive-order samplers fit this view directly, as each one specifies the accept-stay-noise point for every position at every step (Table~\ref{tab:samplers}).
A related split appears in \citet{liu2025think}, which separates planning (choosing which positions to change) from denoising (predicting the clean token).
Our addition is the second axis: prior routing views describe where each token comes from, but not when a position is committed to the KV cache.
The shared notation allows acceptance rules to be compared across corruption processes, while keeping their different noise distributions and constraints on future revisions explicit.

\begin{figure}[tbp]
    \centering
    \newlength{\frontierheight}\setlength{\frontierheight}{0.322\textwidth}%
    \begin{subfigure}[b]{0.47\textwidth}
        \centering
        \includegraphics[height=\frontierheight]{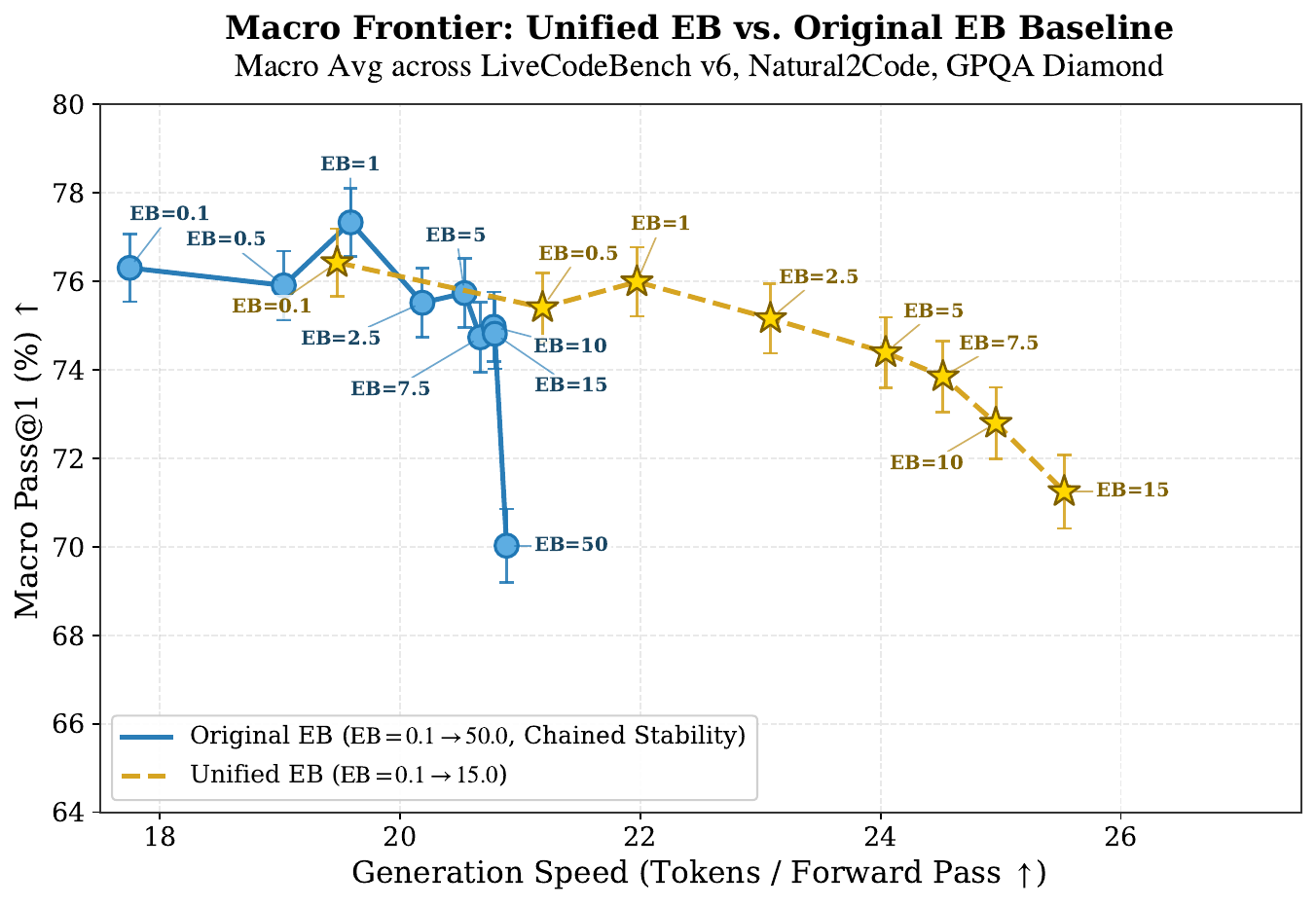}
        \caption{Original vs.\ unified stopping.}
        \label{fig:unified-vs-orig}
    \end{subfigure}
    \hfill
    \begin{subfigure}[b]{0.51\textwidth}
        \centering
        \includegraphics[height=\frontierheight]{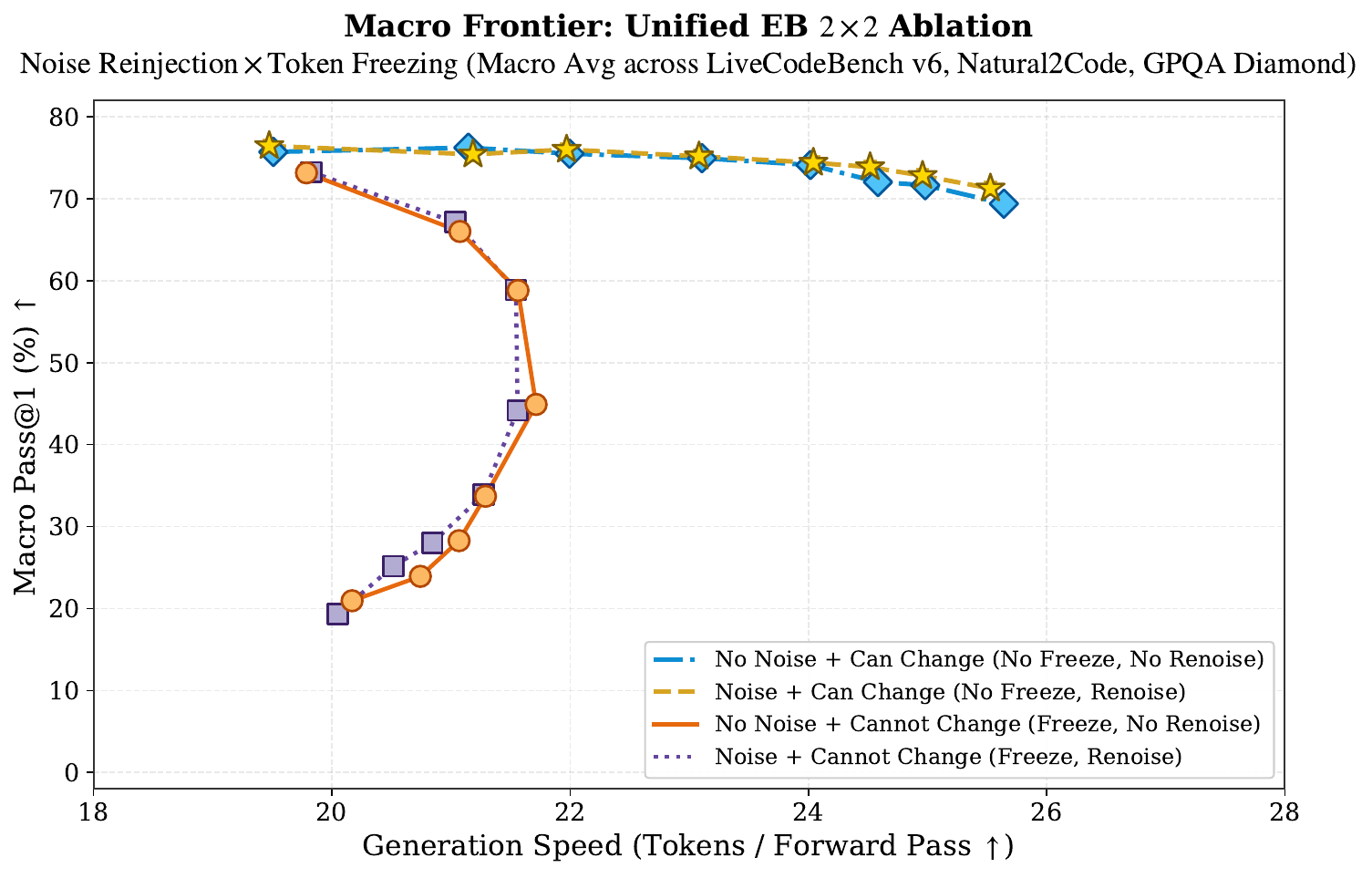}
        \caption{Frozen vs.\ base EB sampler.}
        \label{fig:frozen-eb-frontier}
    \end{subfigure}
    \caption{\textbf{(a)} The unified stopping criterion matches the original at a slight speedup. \textbf{(b)} Simulation of per-position commitment in DiffusionGemma: freezing each token upon acceptance removes the chance for later revision. Frozen variants degrade much faster than unfrozen variants as the entropy bound is relaxed, showing the cost of losing this sampler flexibility.}
    \label{fig:eb-frontiers}
\end{figure}

\paragraph{Baseline: Unified EB}
The default DiffusionGemma sampler accepts tokens with the entropy bound but commits a block with a separate stopping rule (Table~\ref{tab:samplers}).
We replace that rule with the entropy bound itself: the block is committed once every position in the window is accepted in a single step.
This leaves one hyper-parameter, $\gamma$, in control of both choices.
Figure~\ref{fig:unified-vs-orig} shows that the unified rule matches the original rule at the default bound and keeps gaining throughput as $\gamma$ is relaxed, where the original rule stalls.
We use Unified EB as the baseline for the rest of the paper.

\section{Per-Position Commitment: Losing the Chance for Revision}
\label{sec:too-early}
Committing tokens on a per-position basis removes the chance for later revision.
We simulate this consequence of per-position commitment by freezing each token as soon as it is accepted by the EB rule.
The main property we study is that the token can no longer change.
We find that later revision becomes increasingly useful when trading off performance for speed at higher entropy bounds, and that self-correction persists without renoising.

\paragraph{Setup}
We focus on the EB sampler, as it is the default sampler used in \citet{team2026diffusiongemma}.
Our goal is to study the loss of sampler flexibility under per-position commitment, and determine whether later revisions require explicit noise injection.
To do this, we measure the macro-average of LCB v6, Natural2Code, and GPQA Diamond.
We vary two sampler choices: whether accepted positions are frozen, and whether rejected positions are renoised. The model weights and uniform corruption process are unchanged in all four conditions.

\paragraph{Frozen EB Sampler}
The Frozen EB ablation simulates per-position commitment: once a position is accepted under the EB rule, its token can no longer change. We retain the denoising window and blockwise cache commitment to isolate the loss of later revision. This simulation does not reproduce the attention changes or computational savings of actually moving individual positions into the KV cache.
We refer to the unmodified Unified EB sampler as ``unconstrained'', as any position can be changed regardless of whether it was already accepted under the EB rule.

\paragraph{No Noise}
To test whether revisions require explicit noise injection, we additionally disable renoising of rejected positions.
In this condition, rejected positions retain their current tokens, while accepted predictions can still revise unfrozen positions. This isolates the role of explicit renoising without changing the underlying model.

\paragraph{Ablation}
Removing noise and freezing accepted tokens are two independent interventions. Thus we ablate over all four combinations and show the results in Figure \ref{fig:frozen-eb-frontier}.
For each possible combination, we do a sweep over various entropy bound values to examine how they are affected by more relaxed thresholds.
For the unconstrained variants, we also track the empirical frequencies of each position being altered during the denoising trajectory after the initial acceptance under the EB rule.
We show these position volatility metrics in Figure~\ref{fig:wasted} (bottom row).

\paragraph{Results}
For each adaptive variant, the unconstrained sampler achieves a better Pareto frontier than the frozen sampler, and the gap grows as the entropy bound is relaxed.
At the loosest bound, the frozen variants fall to about 20\% macro Pass@1, while the unconstrained variants stay near 70\%.

The number of positions that change after initial acceptance also increases with higher entropy bounds, and is relatively unaffected by whether the sampler uses noise injection.
At the default bound of $0.1$, roughly 97\% to 99\% of accepted positions never change, so the default sampler behaves much like a frozen one; revisions become frequent only as the bound is relaxed.
The simulation shows the cost of removing later revision on a per-position basis: accepting a token once does not mean that fixing it will preserve performance at higher entropy bounds. Self-correction also persists without renoising, so explicit noise injection is not necessary for these revisions.

Per-position commitment therefore loses sampler flexibility as the entropy bound is relaxed. This motivates waiting for more than an individual position to be accepted before removing its chance for later revision.

\section{Block-Level Commitment: Wasting Denoising Slots}
\label{sec:too-late}
At the other extreme, commitment is made at a block level: Unified EB retains every position in the denoising window until the whole block satisfies its stopping rule.
Here we examine the order in which DiffusionGemma accepts tokens under the Unified EB sampler.
DiffusionGemma accepts tokens in an order strongly biased toward autoregressive, left-to-right generation.
This autoregressive bias results in many denoising slots within a canvas being "wasted" on tokens that have already converged.
As a result, a form of rolling sampler where the window moves incrementally through the sequence rather than block by block becomes increasingly appealing as a way to extend the Pareto frontier.

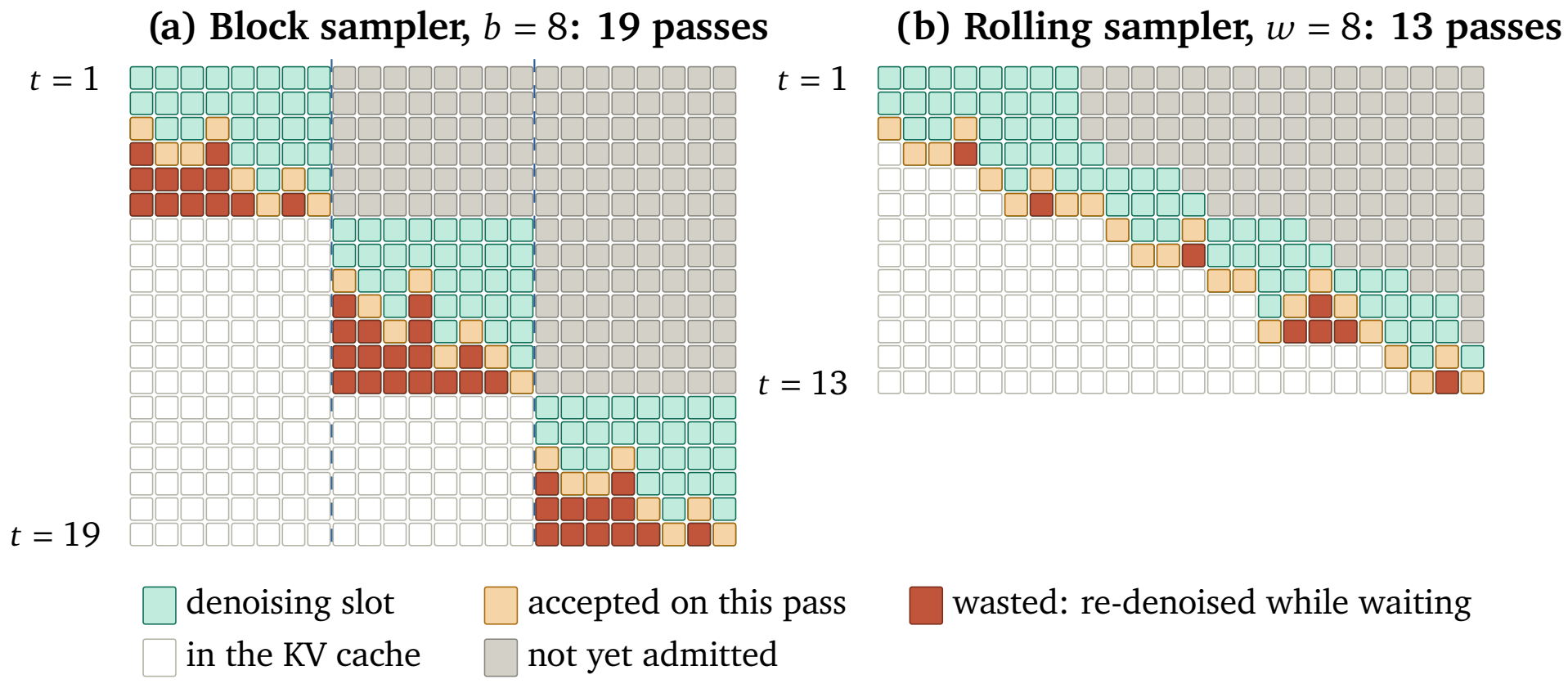
\begin{figure}[tbp]
\centering
\begin{tikzpicture}[x=0.205cm, y=0.205cm]

\definecolor{cwinfill}{HTML}{9FE1CB}
\definecolor{cwinedge}{HTML}{0F6E56}
\definecolor{camber}{HTML}{E8A13C}
\definecolor{camberedge}{HTML}{A96A0B}
\definecolor{cwaste}{HTML}{C0553C}
\definecolor{cwasteedge}{HTML}{7A2E1C}
\definecolor{cnoise}{HTML}{D3D1C7}
\definecolor{cnoiseedge}{HTML}{888780}
\definecolor{cclean}{HTML}{FFFFFF}
\definecolor{ccleanedge}{HTML}{B4B4A9}
\definecolor{cframe}{HTML}{2B5E94}

\newcommand{\gridcell}[3]{%
  \ifcase#3\relax
  \or \fill[cwinfill!65, draw=cwinedge,   line width=0.3pt, rounded corners=0.5pt] (#1, -#2) rectangle ++(0.88, -0.88);
  \or \fill[camber!45,   draw=camberedge, line width=0.3pt, rounded corners=0.5pt] (#1, -#2) rectangle ++(0.88, -0.88);
  \or \fill[cwaste,      draw=cwasteedge, line width=0.3pt, rounded corners=0.5pt] (#1, -#2) rectangle ++(0.88, -0.88);
  \or \fill[cclean,      draw=ccleanedge, line width=0.3pt, rounded corners=0.5pt] (#1, -#2) rectangle ++(0.88, -0.88);
  \or \fill[cnoise,      draw=cnoiseedge, line width=0.3pt, rounded corners=0.5pt] (#1, -#2) rectangle ++(0.88, -0.88);
  \fi}

\newcommand{\gridrow}[4]{%
  \foreach \c in {0,...,23} {%
    \pgfmathtruncatemacro{\sty}{\c<#3 ? 4 : (\c<#4 ? 1 : 5)}%
    \pgfmathsetmacro{\xpos}{\c + #1}%
    \gridcell{\xpos}{#2}{\sty}%
  }}

\newcommand{\gridmarks}[4]{%
  \foreach \c in {#4} {%
    \pgfmathsetmacro{\xpos}{\c + #1}%
    \gridcell{\xpos}{#2}{#3}%
  }}

\newcommand{\keydenoise}{\begin{tikzpicture}[baseline=-0.5ex, x=0.30cm, y=0.30cm]
  \fill[cwinfill!65, draw=cwinedge, line width=0.3pt, rounded corners=0.5pt] (0,-0.5) rectangle ++(0.88,1);
\end{tikzpicture}}
\newcommand{\keyaccept}{\begin{tikzpicture}[baseline=-0.5ex, x=0.30cm, y=0.30cm]
  \fill[camber!45, draw=camberedge, line width=0.3pt, rounded corners=0.5pt] (0,-0.5) rectangle ++(0.88,1);
\end{tikzpicture}}
\newcommand{\keywaste}{\begin{tikzpicture}[baseline=-0.5ex, x=0.30cm, y=0.30cm]
  \fill[cwaste, draw=cwasteedge, line width=0.3pt, rounded corners=0.5pt] (0,-0.5) rectangle ++(0.88,1);
\end{tikzpicture}}
\newcommand{\keycache}{\begin{tikzpicture}[baseline=-0.5ex, x=0.30cm, y=0.30cm]
  \fill[cclean, draw=ccleanedge, line width=0.3pt, rounded corners=0.5pt] (0,-0.5) rectangle ++(0.88,1);
\end{tikzpicture}}
\newcommand{\keyunadmitted}{\begin{tikzpicture}[baseline=-0.5ex, x=0.30cm, y=0.30cm]
  \fill[cnoise, draw=cnoiseedge, line width=0.3pt, rounded corners=0.5pt] (0,-0.5) rectangle ++(0.88,1);
\end{tikzpicture}}

\node[anchor=west, font=\footnotesize\bfseries] at (0, 1.5) {(a) Block sampler, $b=8$: 19 passes};

\gridrow{0}{0}{0}{8}
\gridrow{0}{1}{0}{8}
\gridrow{0}{2}{0}{8}                     \gridmarks{0}{2}{2}{0,3}
\gridrow{0}{3}{0}{8} \gridmarks{0}{3}{3}{0,3}          \gridmarks{0}{3}{2}{1,2}
\gridrow{0}{4}{0}{8} \gridmarks{0}{4}{3}{0,1,2,3}      \gridmarks{0}{4}{2}{4,6}
\gridrow{0}{5}{0}{8} \gridmarks{0}{5}{3}{0,1,2,3,4,6}  \gridmarks{0}{5}{2}{5,7}
\gridrow{0}{6}{8}{16}
\gridrow{0}{7}{8}{16}
\gridrow{0}{8}{8}{16}                        \gridmarks{0}{8}{2}{8,11}
\gridrow{0}{9}{8}{16}  \gridmarks{0}{9}{3}{8,11}             \gridmarks{0}{9}{2}{9}
\gridrow{0}{10}{8}{16} \gridmarks{0}{10}{3}{8,9,11}          \gridmarks{0}{10}{2}{10,13}
\gridrow{0}{11}{8}{16} \gridmarks{0}{11}{3}{8,9,10,11,13}    \gridmarks{0}{11}{2}{12,14}
\gridrow{0}{12}{8}{16} \gridmarks{0}{12}{3}{8,9,10,11,12,13,14} \gridmarks{0}{12}{2}{15}
\gridrow{0}{13}{16}{24}
\gridrow{0}{14}{16}{24}
\gridrow{0}{15}{16}{24}                          \gridmarks{0}{15}{2}{16,19}
\gridrow{0}{16}{16}{24} \gridmarks{0}{16}{3}{16,19}              \gridmarks{0}{16}{2}{17,18}
\gridrow{0}{17}{16}{24} \gridmarks{0}{17}{3}{16,17,18,19}        \gridmarks{0}{17}{2}{20,22}
\gridrow{0}{18}{16}{24} \gridmarks{0}{18}{3}{16,17,18,19,20,22}  \gridmarks{0}{18}{2}{21,23}

\foreach \x in {7.94, 15.94} {
  \draw[cframe, dashed, line width=0.4pt] (\x, 0.3) -- (\x, -19.2);
}
\node[anchor=east, font=\scriptsize] at (-0.5, -0.44) {$t=1$};
\node[anchor=east, font=\scriptsize] at (-0.5, -18.44) {$t=19$};

\node[anchor=west, font=\footnotesize\bfseries] at (29.5, 1.5) {(b) Rolling sampler, $w=8$: 13 passes};

\gridrow{29.5}{0}{0}{8}
\gridrow{29.5}{1}{0}{8}
\gridrow{29.5}{2}{0}{8}                          \gridmarks{29.5}{2}{2}{0,3}
\gridrow{29.5}{3}{1}{9}   \gridmarks{29.5}{3}{3}{3}          \gridmarks{29.5}{3}{2}{1,2}
\gridrow{29.5}{4}{4}{12}                         \gridmarks{29.5}{4}{2}{4,6}
\gridrow{29.5}{5}{5}{13}  \gridmarks{29.5}{5}{3}{6}          \gridmarks{29.5}{5}{2}{5,7,8}
\gridrow{29.5}{6}{9}{17}                         \gridmarks{29.5}{6}{2}{9,12}
\gridrow{29.5}{7}{10}{18} \gridmarks{29.5}{7}{3}{12}         \gridmarks{29.5}{7}{2}{10,11}
\gridrow{29.5}{8}{13}{21}                        \gridmarks{29.5}{8}{2}{13,14,17}
\gridrow{29.5}{9}{15}{23} \gridmarks{29.5}{9}{3}{17}         \gridmarks{29.5}{9}{2}{16,18}
\gridrow{29.5}{10}{15}{23} \gridmarks{29.5}{10}{3}{16,17,18} \gridmarks{29.5}{10}{2}{15,19}
\gridrow{29.5}{11}{20}{24}                       \gridmarks{29.5}{11}{2}{20,22}
\gridrow{29.5}{12}{21}{24} \gridmarks{29.5}{12}{3}{22}       \gridmarks{29.5}{12}{2}{21,23}

\node[anchor=east, font=\scriptsize] at (29.0, -0.44) {$t=1$};
\node[anchor=east, font=\scriptsize] at (29.0, -12.44) {$t=13$};

\node[anchor=north, font=\scriptsize, inner sep=0pt] at (26.7, -20.4) {%
  \begin{tabular}{@{}l@{~}l@{\hspace{1.8em}}l@{~}l@{\hspace{1.8em}}l@{~}l@{}}
    \keydenoise & denoising slot   & \keyaccept     & accepted on this pass &
    \keywaste   & wasted: re-denoised while waiting \\[2.5pt]
    \keycache   & in the KV cache  & \keyunadmitted & not yet admitted      &
                &                                    \\
  \end{tabular}};

\end{tikzpicture}
\caption{\textbf{Where the denoising slots go.} Both samplers denoise a
width-$8$ window per pass in the same order, and the colours are those of
Figure~\ref{fig:algorithm-concept}: teal = still denoising, amber = accepted
on this pass, white = already encoded and in the KV cache, gray = not yet
admitted.  Brick marks a slot spent re-denoising a position that was already
accepted but is still waiting for its prefix, that is, wasted work.
\textbf{(a)} The block sampler appends only once the block is fully resolved,
so converged positions idle and each block re-pays warm-up.  \textbf{(b)} The
rolling sampler appends the accepted prefix immediately and admits fresh
positions in its place, wasting far fewer slots.}
\label{fig:block-vs-rolling}
\end{figure}

\paragraph{Defining Redundant Computations}
Here we define what constitutes a redundant computation.
Our goal is to measure the computations that could conceivably be omitted without drastically impacting the model's behavior.
Thus we define wasted denoising slots as positions that have already converged and are part of a contiguous prefix that could be appended to the KV cache.
We illustrate this definition in Figure \ref{fig:block-vs-rolling}, against the ideal sampler that commits them to the KV cache and avoids redundant denoising steps.

\begin{figure}[tbp]
    \centering
    \includegraphics[width=\linewidth]{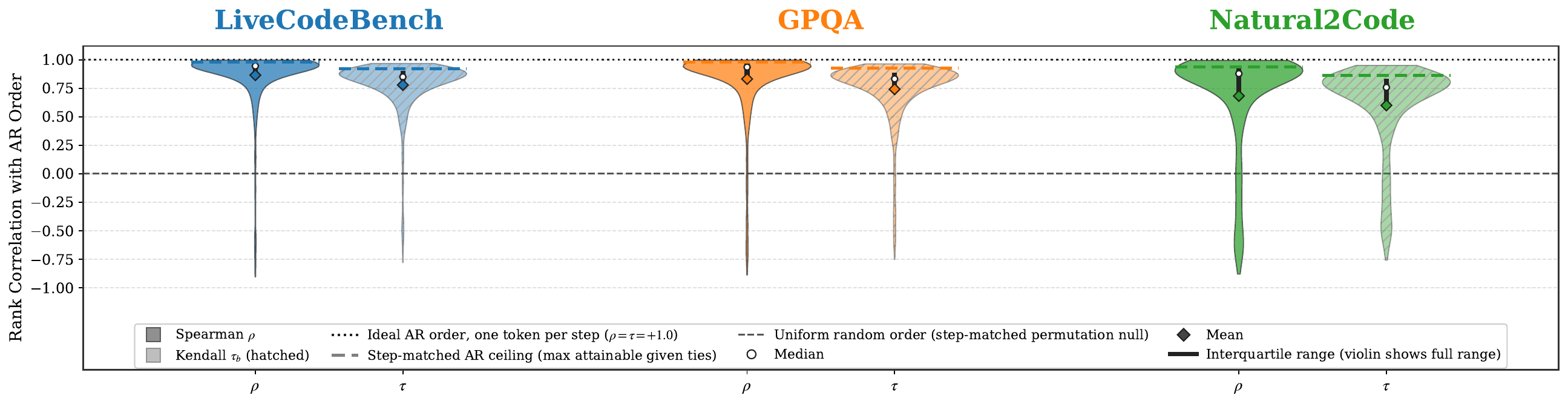}

    \vspace{8pt}

    \includegraphics[width=\linewidth]{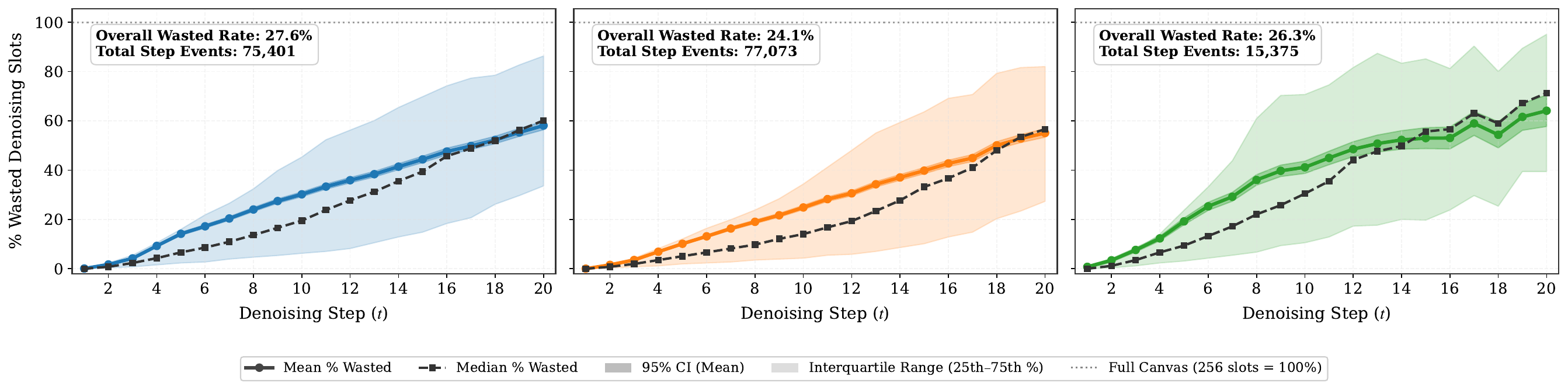}

    \vspace{8pt}

    \includegraphics[width=\linewidth,trim=0 0 0 23.5pt,clip]{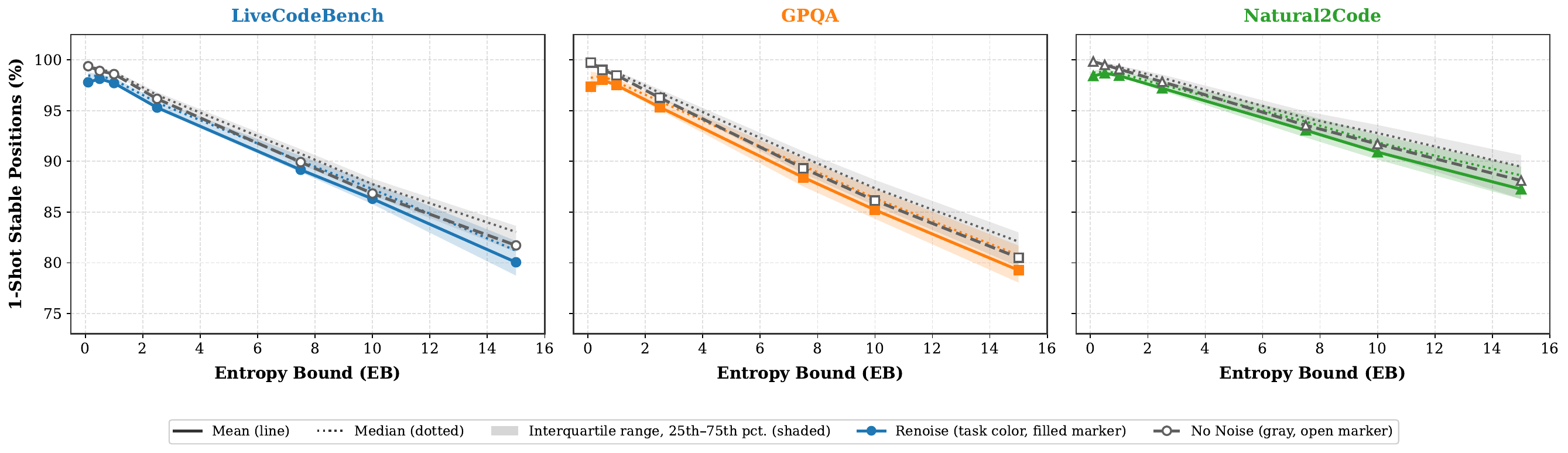}
    \caption{Columns in the top and middle rows show LiveCodeBench, GPQA Diamond, and Natural2Code. \textbf{Top row:} Spearman $\rho$ and Kendall $\tau$ between the acceptance order of the DiffusionGemma EB sampler and the AR generation order; across all tasks the sampler is strongly biased towards left-to-right order. \textbf{Middle row:} Percentage of wasted denoising slots averaged across canvases for LCB v6, GPQA Diamond and Natural2Code. For a given sampling step, a slot is wasted if it belongs to a contiguous prefix that has already converged to its final token. By denoising step 20, the mean fraction of wasted canvas slots exceeds 50\% on all three tasks. \textbf{Bottom row:} Accepted tokens are revised more often at higher entropy bounds (in both argmax and sampled-token stability), independent of renoising.}
    \label{fig:wasted}
\end{figure}

\paragraph{Experimental Setup}
We run the Unified EB sampler on LCB v6, GPQA Diamond and Natural2Code. We track the trajectories of each denoising block, and consider a position as converged when it realizes the value that ends up being committed to the KV cache.

\paragraph{Results}
In Figure~\ref{fig:wasted} (top row), we observe that the first-acceptance order is strongly correlated with left-to-right order (Appendix~\ref{app:order-stats}). At the default entropy bound, accepted positions rarely change, so this ordering serves as a proxy for convergence order. This is consistent with the behavior of wasted denoising slots in Figure~\ref{fig:wasted} (middle row), where we observe that the percentage of wasted slots generally increases during the denoising trajectory.

\citet{asaria2026neither} previously observed this autoregressive bias in the DiffusionGemma sampler.
This has been observed as a general pattern for confidence-based samplers across larger diffusion language models \citep{kim2026interface, gong2026diffucoder, lu2026adablock, yao2026decoding}.
\citet{kim2026interface} demonstrates that confidence based samplers tend to generate samples that are very similar to those produced by left-to-right autoregressive order.
Other works demonstrate that the ordering itself is highly left to right and provide various explanations for why \citep{gong2026diffucoder, lu2026adablock, yao2026decoding}.

Convergence in this measurement is identified retrospectively; during sampling, we use EB acceptance as the operational criterion for deciding which positions are ready to commit.
Where per-position commitment loses the chance for later revision, block-level commitment tolerates redundant computation on converged prefixes. Section~\ref{sec:method} introduces RES to strike a balance: wait for a contiguous accepted prefix, without waiting for the entire denoising window.

\section{RES: A Rolling Sampler}
\label{sec:method}
\begin{figure}[tbp]
\centering
\begin{tikzpicture}[x=0.36cm, y=0.36cm]

\definecolor{cclean}{HTML}{FFFFFF}
\definecolor{ccleanedge}{HTML}{B4B4A9}
\definecolor{cwinedge}{HTML}{0F6E56}
\definecolor{cwinfill}{HTML}{9FE1CB}
\definecolor{cnoise}{HTML}{D3D1C7}
\definecolor{cnoiseedge}{HTML}{888780}
\definecolor{cframe}{HTML}{2B5E94}
\definecolor{camber}{HTML}{E8A13C}
\definecolor{camberedge}{HTML}{A96A0B}
\definecolor{mempty}{HTML}{F5F4EF}
\definecolor{memptyedge}{HTML}{C9C7BE}

\newcommand{\cellclean}[3]{%
  \fill[cclean, draw=ccleanedge, line width=0.4pt, rounded corners=1pt] (#1, #2) rectangle ++(0.9, 1);
  \node[font=\scriptsize\ttfamily, inner sep=0pt, text=black] at (#1+0.45, #2+0.5) {#3};}
\newcommand{\cellwin}[4]{%
  \fill[cwinfill!#3, draw=cwinedge, line width=0.4pt, rounded corners=1pt] (#1, #2) rectangle ++(0.9, 1);
  \node[font=\scriptsize\ttfamily, inner sep=0pt, text=black] at (#1+0.45, #2+0.5) {#4};}
\newcommand{\cellnoise}[3]{%
  \fill[cnoise, draw=cnoiseedge, line width=0.4pt, rounded corners=1pt] (#1, #2) rectangle ++(0.9, 1);
  \node[font=\scriptsize\ttfamily, inner sep=0pt, text=cnoiseedge] at (#1+0.45, #2+0.5) {#3};}
\newcommand{\cellamber}[3]{%
  \fill[camber!45, draw=camberedge, line width=0.6pt, rounded corners=1pt] (#1, #2) rectangle ++(0.9, 1);
  \node[font=\scriptsize\ttfamily, inner sep=0pt, text=black] at (#1+0.45, #2+0.5) {#3};}
\newcommand{\acceptmark}[2]{\draw[camberedge, line width=1.1pt, rounded corners=1pt] (#1+0.03, #2+0.03) rectangle ++(0.84, 0.94);}
\newcommand{\acceptdot}[2]{\fill[camberedge] (#1+0.45, #2+0.17) circle (0.12);}
\newcommand{\passframe}[2]{%
  \draw[cframe, line width=1.1pt, rounded corners=2pt, dash pattern=on 4pt off 2pt]
    (#1-0.15, #2-0.25) rectangle (#1+10.05, #2+1.25);}
\newcommand{\seqbracket}[4]{%
  \draw[line width=0.3pt] (#1, #3-0.55) -- (#1, #3-0.90) -- (#2+0.9, #3-0.90) -- (#2+0.9, #3-0.55);
  \node[font=\tiny, anchor=north, inner sep=1.5pt] at ({(#1+#2+0.9)/2}, #3-0.90) {#4};}

\newcommand{\maskpanel}[5]{%
\begin{scope}[shift={(#4cm,#5cm)}, x=0.19cm, y=0.19cm]
  \foreach \q in {#1,...,#2}{%
    \foreach \k in {#1,...,#2}{%
      \pgfmathtruncatemacro{\ii}{\k-#1}%
      \pgfmathtruncatemacro{\jj}{\q-#1}%
      \ifnum\q>#3\relax
        \ifnum\k>#3\relax
          \fill[cwinfill!65, draw=cwinedge, line width=0.25pt, rounded corners=0.4pt] (\ii,-\jj) rectangle ++(0.88,-0.88);
        \else
          \fill[camber!20, draw=camberedge, line width=0.25pt, rounded corners=0.4pt] (\ii,-\jj) rectangle ++(0.88,-0.88);
        \fi
      \else
        \ifnum\k>\q\relax
          \fill[mempty, draw=memptyedge, line width=0.25pt, rounded corners=0.4pt] (\ii,-\jj) rectangle ++(0.88,-0.88);
        \else
          \fill[camber!55, draw=camberedge, line width=0.25pt, rounded corners=0.4pt] (\ii,-\jj) rectangle ++(0.88,-0.88);
        \fi
      \fi}}%
  \pgfmathtruncatemacro{\cc}{#3-#1+1}%
  \pgfmathtruncatemacro{\nn}{#2-#1+1}%
  \ifnum\cc>0\relax
    \draw[dashed, line width=0.4pt] (\cc-0.06, 0.5) -- (\cc-0.06, -\nn-0.1);
    \draw[dashed, line width=0.4pt] (-0.5, -\cc+0.06) -- (\nn+0.1, -\cc+0.06);
    \draw[camberedge, line width=0.8pt, rounded corners=0.6pt] (-0.06, 0.06) rectangle (\cc-0.12, -\cc+0.12);
  \fi
\end{scope}}
\newcommand{\maskbrackets}[4]{%
\begin{scope}[shift={(#3cm,#4cm)}, x=0.19cm, y=0.19cm]
  \ifnum#1>0\relax
    \draw[line width=0.3pt] (0,-#2-0.5) -- (0,-#2-0.9) -- (#1-0.12,-#2-0.9) -- (#1-0.12,-#2-0.5);
    \node[font=\tiny, anchor=north, inner sep=1.5pt] at (#1/2-0.06, -#2-0.9) {prefix};
  \fi
  \draw[line width=0.3pt] (#1,-#2-0.5) -- (#1,-#2-0.9) -- (#2-0.12,-#2-0.9) -- (#2-0.12,-#2-0.5);
  \node[font=\tiny, anchor=north, inner sep=1.5pt] at ({(#1+#2)/2-0.06}, -#2-0.9) {window};
\end{scope}}

\newcommand{\keyencoded}{\begin{tikzpicture}[baseline=-0.5ex, x=0.36cm, y=0.36cm]
  \fill[cclean, draw=ccleanedge, line width=0.4pt, rounded corners=1pt] (0,-0.5) rectangle ++(0.9,1);
\end{tikzpicture}}
\newcommand{\keyaccepted}{\begin{tikzpicture}[baseline=-0.5ex, x=0.36cm, y=0.36cm]
  \fill[cwinfill!30, draw=cwinedge, line width=0.4pt, rounded corners=1pt] (0,-0.5) rectangle ++(0.9,1);
  \draw[camberedge, line width=1.1pt, rounded corners=1pt] (0.03,-0.47) rectangle ++(0.84,0.94);
\end{tikzpicture}}
\newcommand{\keywaiting}{\begin{tikzpicture}[baseline=-0.5ex, x=0.36cm, y=0.36cm]
  \fill[cwinfill!30, draw=cwinedge, line width=0.4pt, rounded corners=1pt] (0,-0.5) rectangle ++(0.9,1);
  \fill[camberedge] (0.45,-0.33) circle (0.12);
\end{tikzpicture}}
\newcommand{\keyprefix}{\begin{tikzpicture}[baseline=-0.5ex, x=0.36cm, y=0.36cm]
  \fill[camber!45, draw=camberedge, line width=0.6pt, rounded corners=1pt] (0,-0.5) rectangle ++(0.9,1);
\end{tikzpicture}}
\newcommand{\keynoise}{\begin{tikzpicture}[baseline=-0.5ex, x=0.36cm, y=0.36cm]
  \fill[cnoise, draw=cnoiseedge, line width=0.4pt, rounded corners=1pt] (0,-0.5) rectangle ++(0.9,1);
  \node[font=\scriptsize\ttfamily, inner sep=0pt, text=cnoiseedge] at (0.45,0) {k};
\end{tikzpicture}}
\newcommand{\keymcausal}{\begin{tikzpicture}[baseline=-0.5ex, x=0.36cm, y=0.36cm]
  \fill[camber!55, draw=camberedge, line width=0.4pt, rounded corners=1pt] (0,-0.5) rectangle ++(0.9,1);
\end{tikzpicture}}
\newcommand{\keymcross}{\begin{tikzpicture}[baseline=-0.5ex, x=0.36cm, y=0.36cm]
  \fill[camber!20, draw=camberedge, line width=0.4pt, rounded corners=1pt] (0,-0.5) rectangle ++(0.9,1);
\end{tikzpicture}}
\newcommand{\keymbidi}{\begin{tikzpicture}[baseline=-0.5ex, x=0.36cm, y=0.36cm]
  \fill[cwinfill!65, draw=cwinedge, line width=0.4pt, rounded corners=1pt] (0,-0.5) rectangle ++(0.9,1);
\end{tikzpicture}}
\newcommand{\keymempty}{\begin{tikzpicture}[baseline=-0.5ex, x=0.36cm, y=0.36cm]
  \fill[mempty, draw=memptyedge, line width=0.4pt, rounded corners=1pt] (0,-0.5) rectangle ++(0.9,1);
\end{tikzpicture}}

\node[anchor=east, font=\small\bfseries] at (-0.60, 0.50) {step $t$};
\cellclean{0}{0}{t}
\cellclean{1}{0}{h}
\cellclean{2}{0}{e}
\cellclean{3}{0}{}
\cellclean{4}{0}{c}
\cellclean{5}{0}{a}
\cellclean{6}{0}{t}
\cellclean{7}{0}{}
\cellwin{8}{0}{30}{m}
\cellwin{9}{0}{30}{e}
\cellwin{10}{0}{55}{r}
\cellwin{11}{0}{30}{w}
\cellwin{12}{0}{70}{s}
\cellwin{13}{0}{90}{}
\cellwin{14}{0}{100}{q}
\cellwin{15}{0}{100}{7}
\cellwin{16}{0}{100}{x}
\cellwin{17}{0}{100}{v}
\cellnoise{18}{0}{9}
\cellnoise{19}{0}{k}
\cellnoise{20}{0}{z}
\cellnoise{21}{0}{3}
\cellnoise{22}{0}{b}
\cellnoise{23}{0}{5}
\acceptmark{8}{0}
\acceptmark{9}{0}
\acceptmark{11}{0}
\passframe{8}{0}
\seqbracket{8}{17}{0}{10 positions per pass}
\maskpanel{8}{17}{7}{9.25}{1.13}

\node[font=\scriptsize, anchor=south, text=ccleanedge!60!black] at (3.95, 1.32) {KV cache};
\node[cframe, font=\scriptsize, anchor=south] at (12.95, 1.32) {processed this pass};
\node[cnoiseedge, font=\scriptsize, anchor=south] at (20.95, 1.32) {not yet admitted};
\node[font=\scriptsize, anchor=south] at (10.20cm, 1.57cm) {attention mask};
\node[font=\tiny, anchor=south] at (10.20cm, 1.19cm) {keys $\rightarrow$};
\node[font=\tiny, rotate=90, anchor=south] at (9.09cm, 0.18cm) {queries};

\node[anchor=east, font=\small\bfseries] at (-0.60, -6.50) {step $t{+}1$};
\cellclean{0}{-7.00}{t}
\cellclean{1}{-7.00}{h}
\cellclean{2}{-7.00}{e}
\cellclean{3}{-7.00}{}
\cellclean{4}{-7.00}{c}
\cellclean{5}{-7.00}{a}
\cellclean{6}{-7.00}{t}
\cellclean{7}{-7.00}{}
\cellamber{8}{-7.00}{m}
\cellamber{9}{-7.00}{e}
\cellwin{10}{-7.00}{30}{o}
\cellwin{11}{-7.00}{30}{w}
\cellwin{12}{-7.00}{30}{s}
\cellwin{13}{-7.00}{30}{}
\cellwin{14}{-7.00}{55}{l}
\cellwin{15}{-7.00}{55}{o}
\cellwin{16}{-7.00}{70}{u}
\cellwin{17}{-7.00}{70}{d}
\cellnoise{18}{-7.00}{9}
\cellnoise{19}{-7.00}{k}
\cellnoise{20}{-7.00}{z}
\cellnoise{21}{-7.00}{3}
\cellnoise{22}{-7.00}{b}
\cellnoise{23}{-7.00}{5}
\acceptmark{10}{-7.00}
\acceptdot{11}{-7.00}
\acceptmark{12}{-7.00}
\acceptmark{13}{-7.00}
\passframe{8}{-7.00}
\maskpanel{8}{17}{9}{9.25}{-1.39}

\node[anchor=east, font=\small\bfseries] at (-0.60, -13.50) {step $t{+}2$};
\cellclean{0}{-14.00}{t}
\cellclean{1}{-14.00}{h}
\cellclean{2}{-14.00}{e}
\cellclean{3}{-14.00}{}
\cellclean{4}{-14.00}{c}
\cellclean{5}{-14.00}{a}
\cellclean{6}{-14.00}{t}
\cellclean{7}{-14.00}{}
\cellclean{8}{-14.00}{m}
\cellclean{9}{-14.00}{e}
\cellamber{10}{-14.00}{o}
\cellamber{11}{-14.00}{w}
\cellamber{12}{-14.00}{s}
\cellamber{13}{-14.00}{}
\cellwin{14}{-14.00}{30}{l}
\cellwin{15}{-14.00}{30}{o}
\cellwin{16}{-14.00}{30}{u}
\cellwin{17}{-14.00}{30}{d}
\cellwin{18}{-14.00}{100}{9}
\cellwin{19}{-14.00}{100}{k}
\cellnoise{20}{-14.00}{z}
\cellnoise{21}{-14.00}{3}
\cellnoise{22}{-14.00}{b}
\cellnoise{23}{-14.00}{5}
\acceptmark{14}{-14.00}
\acceptmark{15}{-14.00}
\acceptmark{16}{-14.00}
\acceptmark{17}{-14.00}
\passframe{10}{-14.00}
\seqbracket{10}{13}{-14.00}{encoded prefix}
\seqbracket{14}{19}{-14.00}{denoising window}
\maskpanel{10}{19}{13}{9.25}{-3.91}
\maskbrackets{4}{10}{9.25}{-3.91}

\node[anchor=north, font=\scriptsize, inner sep=0pt] at (5.10cm, -6.60cm) {%
  \begin{tabular}{@{}l@{~}l@{\hspace{2.2em}}l@{~}l@{}}
    \keyencoded  & already encoded, in the KV cache   & \keymcausal & prefix $\to$ prefix (causal) \\[2.5pt]
    \keyprefix   & committed, encoded on this pass    & \keymcross  & window $\to$ prefix \\[2.5pt]
    \keyaccepted & accepted on this pass              & \keymbidi   & window $\leftrightarrow$ window (bidirectional) \\[2.5pt]
    \keywaiting  & accepted earlier, awaiting prefix  & \keymempty  & masked out \\[2.5pt]
    \keynoise    & not yet admitted (noise token)     &             & \\
  \end{tabular}};

\end{tikzpicture}
\caption{Three passes of the rolling sampler.  Every pass spends a fixed
budget of ten positions (dashed frame), split between an \emph{encoded
prefix} of the tokens committed on the previous pass and the remaining
\emph{denoising window}; the mask panel on the right is exactly that block of
ten.  \textbf{Left:} positions keep their column, so each slot can be read down
its column.  Out-of-order acceptances wait for a contiguous prefix (\texttt{w}
at $t$) and unaccepted slots stay mutable (column ten: \texttt{r} $\to$
\texttt{o}).  The prefix completing at $t{+}1$ is four slots long, so four
slots must be encoded at $t{+}2$: the block advances by however many positions
graduated into the KV cache, and the split is $0{+}10$, then $2{+}8$, then
$4{+}6$.  \textbf{Right:} rows are queries, columns are keys; a coloured cell
means the query attends to the key.  The encoded prefix is causal, so its
KV-cache entries depend only on committed tokens and are never revised, while
denoising positions are bidirectional among themselves and attend fully to the
prefix.  Committing more positions per pass therefore moves the KV cache forward
faster, but leaves fewer positions to denoise on the next pass.  Every query additionally attends to the
whole KV cache (white cells), which the panels omit for space.}
\label{fig:algorithm-concept}
\end{figure}
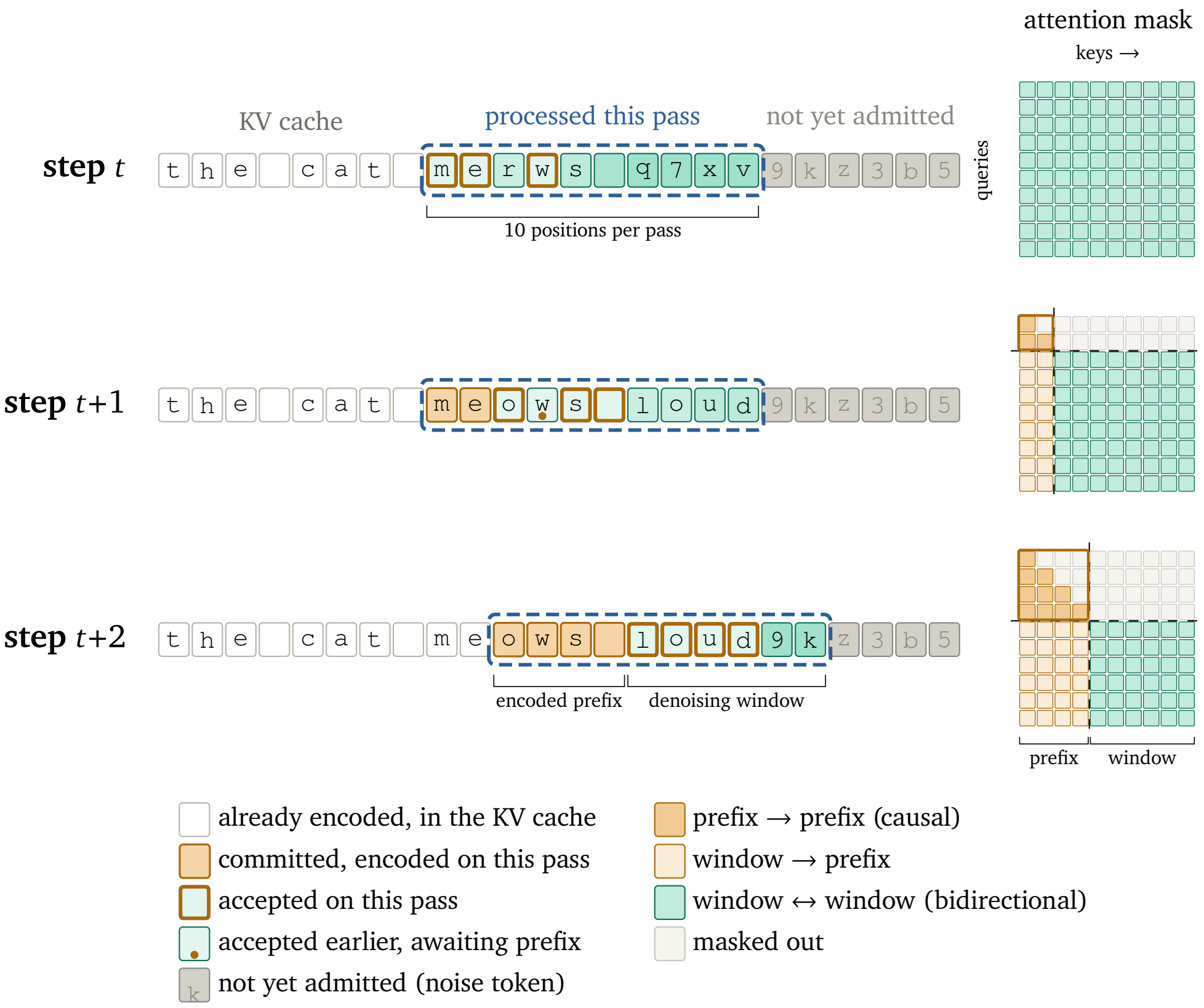

Here we introduce the Rolling Entropy Sampler (RES), a middle ground between per-position commitment and block-level commitment. Rather than committing each token as soon as it is accepted or waiting for the entire window, RES waits until a contiguous prefix of at least $\ell_{\min}$ positions has been accepted by the EB rule.
The core mechanism is simple: instead of waiting for the entire block to converge, RES encodes any sufficiently long contiguous prefix of accepted tokens, appends it to the KV cache, and refills the freed slots with fresh noise.
An accepted position farther ahead remains free to change until the preceding positions are also accepted and the minimum-shift condition is met. Waiting for the contiguous prefix preserves the chance for later revision without retaining the entire denoising window.
Figure~\ref{fig:algorithm-concept} illustrates the method.

\paragraph{Commitment Rule}
In the terms of Section~\ref{sec:motivation}, a rolling sampler is an adaptive sampler with a non-trivial commitment rule.
RES repurposes the acceptance rule for the EB sampler to also serve as the commitment rule: once a sufficiently large contiguous prefix of positions has been accepted under the EB rule, these positions are appended to the KV cache. 
The entropy bound criterion controls the predictive entropies of jointly accepted positions. Under the low-prediction-error assumption discussed in Section~\ref{sec:prelims}, \eqref{eq:eb-bound} approximately bounds the joint dependence error incurred by factorized sampling \citep{ben2026accelerated}.
Thus we use the EB criterion to define $F_{commit}$ as well as $F_{route}$: RES commits the longest prefix of the window whose positions all lie in the accepted set $A_{\gamma}$, provided it meets the minimum shift criterion below (Table~\ref{tab:samplers}).
\begin{align*}
    F_{commit}(X)_i &= \mathbb{1}\!\left[\, i \le \ell(A_\gamma) \,\right] \, \mathbb{1}\!\left[\, \ell(A_\gamma) \ge \ell_{\min} \,\right] \\
    A_\gamma &= \arg\max_{U} |U| \quad \text{s.t.} \quad
    \sum_{i \in U} H\big(P_\theta^i(\cdot \mid X_t)\big) - \max_{i \in U} H\big(P_\theta^i(\cdot \mid X_t)\big) \le \gamma \\
    \ell(A) &= \max \{\, j : \{1, \dots, j\} \subseteq A \,\}
\end{align*}
Here, $A_{\gamma}$ represents the set of accepted positions under the EB rule, given entropy threshold $\gamma$; and $\ell(A)$ is the largest contiguous prefix of accepted positions in $A_{\gamma}$. 
We discuss the minimum shift criterion below. 

\paragraph{Minimum Shift Criterion}
To interpolate smoothly between Unified EB and fully incremental commitment, we introduce a minimum shift criterion: the contiguous prefix must contain at least $\ell_{\min}$ tokens in order to be appended to the KV cache. This gives another axis for trading off performance for speed: by decreasing $\ell_{\min}$, it is possible to accelerate RES without altering the entropy bound.
The minimum shift requires acceptance across a sufficiently long prefix before commitment, while leaving accepted positions farther ahead available for revision.

RES is distinct from doing block diffusion with a smaller block size: while block diffusion denoises and commits each block before moving to the next, RES commits a prefix while retaining the remaining positions in the denoising window.
This can be seen as *warm starting* the positions ahead of the committed prefix, as these positions preserve their denoising progress across shifts and remain available for further revision.

\subsection{Additional Modifications}
RES includes two further design choices, and we also note a property that follows from them. 

First, we use a position-based temperature schedule. Positions toward the left of the window have been denoised for longer than those that have just entered on the right, and the schedule reflects this, thus improving speed. 

Second, we merge the encoding and denoising steps. RES commits short prefixes frequently, so a separate encoding pass for each commit would cost more forward passes than rolling saves; merging makes rolling commitment efficient. All reported TPF numbers count every forward pass for both samplers, including Unified EB’s per-block encoding pass. 

Finally, because the temperature schedule no longer depends on the maximum number of denoising steps, that maximum acts only as a budget. We do not evaluate this property here.
\paragraph{Position-based Temperature Schedule}
One natural consequence of the autoregressive bias of DiffusionGemma is that a token's position in the rolling canvas provides a proxy for how close it is to having converged.
Thus we can modify the standard temperature shaping applied in DiffusionGemma to be based on position rather than noise level.
This accelerates tokens that are close to the prefix, as these are closer to being committed, while allowing for positions in the suffix to explore candidate tokens.

\paragraph{Simultaneous Encoding and Denoising}
One other advantage of RES is that the number of tokens appended to the KV cache is much smaller than the total canvas. This allows for merging both the encoding step and the denoising step present in DiffusionGemma.
More specifically, given a window of $w$ positions and a contiguous prefix of $\ell$ accepted tokens, we use a causal mask for positions $0$ to $\ell-1$, while positions $\ell$ to $w-1$ have bidirectional attention.
This keeps the number of processed positions per forward pass the same between the original DiffusionGemma sampler and RES, while ensuring that every step is performing denoising.
We include a visualization of how the attention mask changes alongside the steps of RES in Figure \ref{fig:algorithm-concept}.

\paragraph{Anytime Sampling}
While Unified EB is highly adaptive, the temperature schedule is tied to the maximum DNS, which is a hyper-parameter that must be set.
Ideally, the maximum DNS would only serve a time-out function to prevent sampler stalls, as opposed to controlling the logit shaping.
One consequence of adopting the position-based temperature schedule is that the logit shaping becomes independent of the maximum number of DNS, making it an \textit{anytime sampler}: the maximum number of denoising steps serves as a budget limit rather than controlling fundamental sampler behavior.

\section{Experiments}
\label{sec:experiments}
\newsavebox{\resultspanels}
\begin{lrbox}{\resultspanels}
\begin{minipage}{\textwidth}
\captionsetup{hypcap=false}
\centering
\begin{minipage}[c]{0.49\textwidth}
\footnotesize
\setlength{\tabcolsep}{2pt}
\renewcommand{\arraystretch}{1.0}
\centering
\captionsetup{font=footnotesize}
\captionof{table}{Comparison of \textbf{Unified EB} vs.\ \textbf{RES} across benchmarks ($\text{EB}=0.1, \text{max\_dns}=48$). Scores report \textbf{Avg@20} $\pm$ 95\% CI (Pass@1 averaged across 20 seeds; 5 seeds for LiveCodeBench). Relative TPF deltas are shown underneath.}
\label{tab:eb_unified_vs_rolling}
\vspace{4pt}
\begin{tabular*}{\linewidth}{@{\extracolsep{\fill}}lcccc@{}}
\toprule
& \multicolumn{2}{c}{\textbf{Unified EB}} & \multicolumn{2}{c}{\textbf{RES}} \\
\cmidrule(lr){2-3} \cmidrule(lr){4-5}
\textbf{Task} & \textbf{Avg@20} & \textbf{TPF} & \textbf{Avg@20} & \textbf{TPF} \\
\midrule
LiveCodeBench & \begin{tabular}[t]{@{}c@{}}65.4\% \\ \scriptsize{($\pm$3.2\%)}\end{tabular} & 18.7 & \begin{tabular}[t]{@{}c@{}}62.6\% \\ \scriptsize{($\pm$3.2\%)}\end{tabular} & \begin{tabular}[t]{@{}c@{}}\textbf{26.3} \\ \scriptsize{(+40.6\%)}\end{tabular} \\
\addlinespace[2pt]
Natural2Code & \begin{tabular}[t]{@{}c@{}}\textbf{93.0\%} \\ \scriptsize{\textbf{($\pm$0.8\%)}}\end{tabular} & 22.9 & \begin{tabular}[t]{@{}c@{}}91.3\% \\ \scriptsize{($\pm$0.9\%)}\end{tabular} & \begin{tabular}[t]{@{}c@{}}\textbf{30.1} \\ \scriptsize{(+31.7\%)}\end{tabular} \\
\addlinespace[2pt]
GPQA Diamond & \begin{tabular}[t]{@{}c@{}}67.5\% \\ \scriptsize{($\pm$1.5\%)}\end{tabular} & 16.6 & \begin{tabular}[t]{@{}c@{}}68.3\% \\ \scriptsize{($\pm$1.4\%)}\end{tabular} & \begin{tabular}[t]{@{}c@{}}\textbf{23.8} \\ \scriptsize{(+43.1\%)}\end{tabular} \\
\addlinespace[2pt]
HiddenMath   & \begin{tabular}[t]{@{}c@{}}\textbf{73.9\%} \\ \scriptsize{\textbf{($\pm$0.9\%)}}\end{tabular} & 22.1 & \begin{tabular}[t]{@{}c@{}}72.6\% \\ \scriptsize{($\pm$0.9\%)}\end{tabular} & \begin{tabular}[t]{@{}c@{}}\textbf{27.2} \\ \scriptsize{(+23.2\%)}\end{tabular} \\
\addlinespace[2pt]
GSM8k        & \begin{tabular}[t]{@{}c@{}}\textbf{92.1\%} \\ \scriptsize{\textbf{($\pm$0.3\%)}}\end{tabular} & 21.7 & \begin{tabular}[t]{@{}c@{}}91.6\% \\ \scriptsize{($\pm$0.3\%)}\end{tabular} & \begin{tabular}[t]{@{}c@{}}\textbf{27.7} \\ \scriptsize{(+28.0\%)}\end{tabular} \\
\addlinespace[2pt]
IFEval       & \begin{tabular}[t]{@{}c@{}}\textbf{93.1\%} \\ \scriptsize{\textbf{($\pm$0.5\%)}}\end{tabular} & 16.8 & \begin{tabular}[t]{@{}c@{}}82.4\% \\ \scriptsize{($\pm$0.7\%)}\end{tabular} & \begin{tabular}[t]{@{}c@{}}\textbf{22.6} \\ \scriptsize{(+34.4\%)}\end{tabular} \\
\addlinespace[2pt]
MMLU         & \begin{tabular}[t]{@{}c@{}}80.3\% \\ \scriptsize{($\pm$0.4\%)}\end{tabular} & 16.2 & \begin{tabular}[t]{@{}c@{}}\textbf{82.8\%} \\ \scriptsize{\textbf{($\pm$0.4\%)}}\end{tabular} & \begin{tabular}[t]{@{}c@{}}\textbf{21.0} \\ \scriptsize{(+30.0\%)}\end{tabular} \\
\midrule
\textbf{Macro Average} & \begin{tabular}[t]{@{}c@{}}\textbf{80.7\%} \\ \scriptsize{\textbf{($\pm$0.5\%)}}\end{tabular} & 19.3 & \begin{tabular}[t]{@{}c@{}}78.8\% \\ \scriptsize{($\pm$0.5\%)}\end{tabular} & \begin{tabular}[t]{@{}c@{}}\textbf{25.5} \\ \scriptsize{\textbf{(+32.5\%)}}\end{tabular} \\
\bottomrule
\end{tabular*}

\end{minipage}\hfill
\begin{minipage}[c]{0.49\textwidth}
\centering
\includegraphics[width=\linewidth]{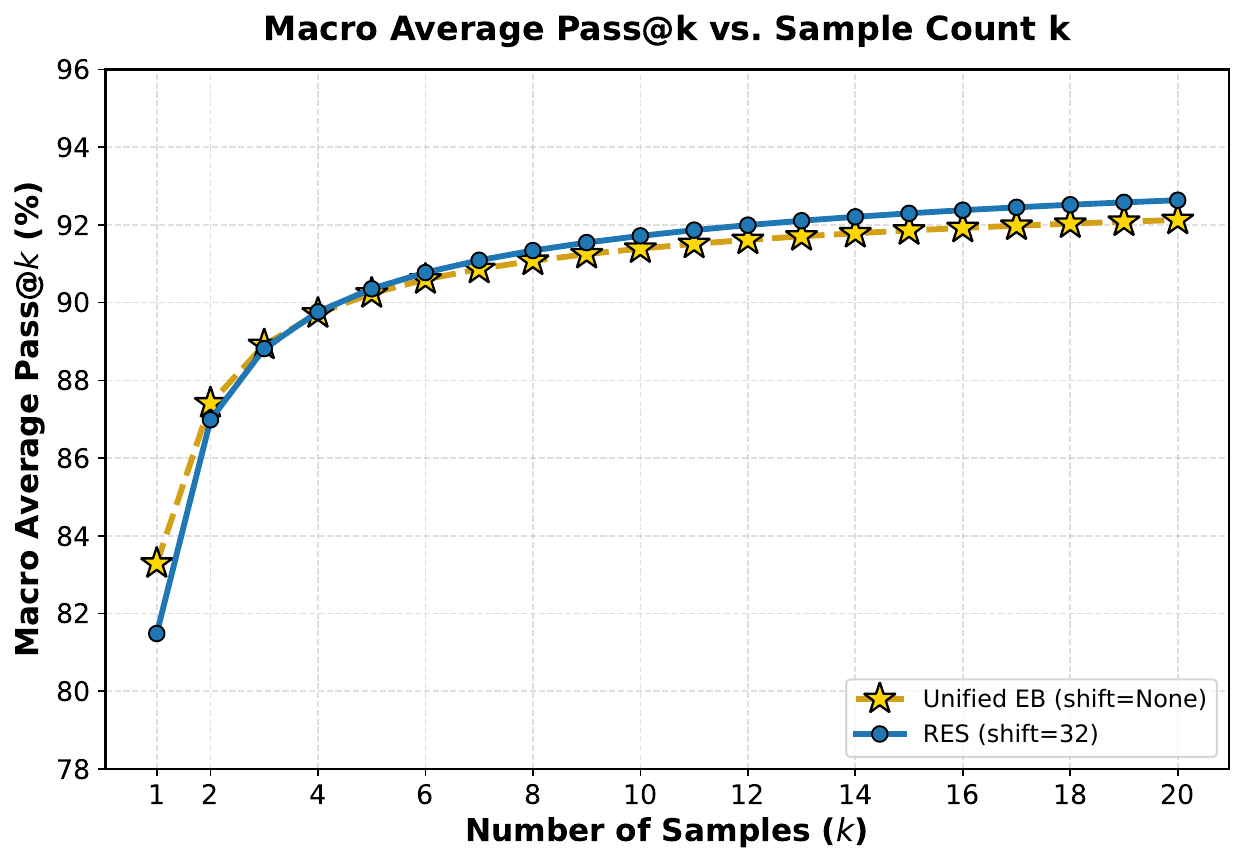}
\captionof{figure}{\textbf{Macro Average Pass@$k$ vs.\ Sample Count $k$}. While Unified EB achieves higher Pass@1, RES ($\ell_{\min}=32$) exhibits stronger multi-sample scaling, overtaking Unified EB at $k \ge 6$ and maintaining its advantage through $k=20$. Table~\ref{tab:eb_unified_vs_rolling} reports a 32.5\% increase in tokens per forward pass.}
\label{fig:macro_pass_at_k}

\end{minipage}
\par\vspace{12pt}
\begin{minipage}{\textwidth}
\centering
    \includegraphics[width=\linewidth]{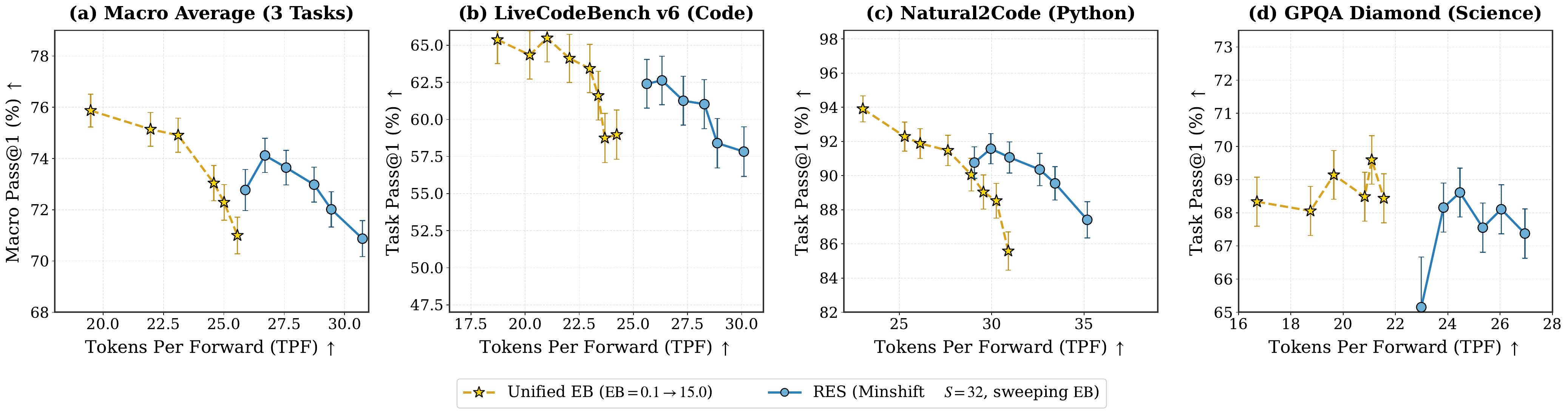}
    \captionof{figure}{Pareto frontiers of RES with minimum shift $\ell_{\min}=32$ and Unified EB under varied entropy bounds for LCB v6 and Natural2Code averaged over 5 seeds, and GPQA Diamond averaged over 20 seeds due to its higher evaluation variance.}
    \label{fig:minshift-16}

\end{minipage}
\end{minipage}
\end{lrbox}

\subsection{Main Results}
We compare RES against Unified EB on seven benchmarks at the default entropy bound $\gamma = 0.1$, averaging over 20 seeds (5 for LiveCodeBench).
Table~\ref{tab:eb_unified_vs_rolling} reports accuracy and tokens per forward pass (TPF).

\begin{figure}[tbp]
\centering
\usebox{\resultspanels}
\end{figure}

\paragraph{Performance and Efficiency}
RES increases TPF on every benchmark, by 23\% to 43\%, and \textbf{by 32.5\% on the macro average} (19.3 to 25.5).
Macro accuracy goes from 80.7\% to 78.8\%.
The cost is uneven across tasks.
Accuracy increases on MMLU (+2.5\% absolute) and is unchanged within confidence intervals on GPQA Diamond.
It falls by 0.5\% to 2.8\% (absolute) on GSM8k, HiddenMath, Natural2Code, and LiveCodeBench, and by 10.7\% on IFEval, the one task where RES is clearly worse.
Excluding IFEval, macro accuracy falls by 0.5\%.
One possible cause of the degradation is exposure bias: RES induces a noise pattern across the window that is heavily left-to-right, which is distinct from the homogeneous noise level used within the block at training time.
We do not test this explanation here.

\subsection{Speed-Quality Frontier}
We next sweep both knobs of RES, using the macro-average across LCB v6, GPQA Diamond, and Natural2Code.

\begin{figure}[tbp]
    \centering
    \includegraphics[width=0.6\linewidth]{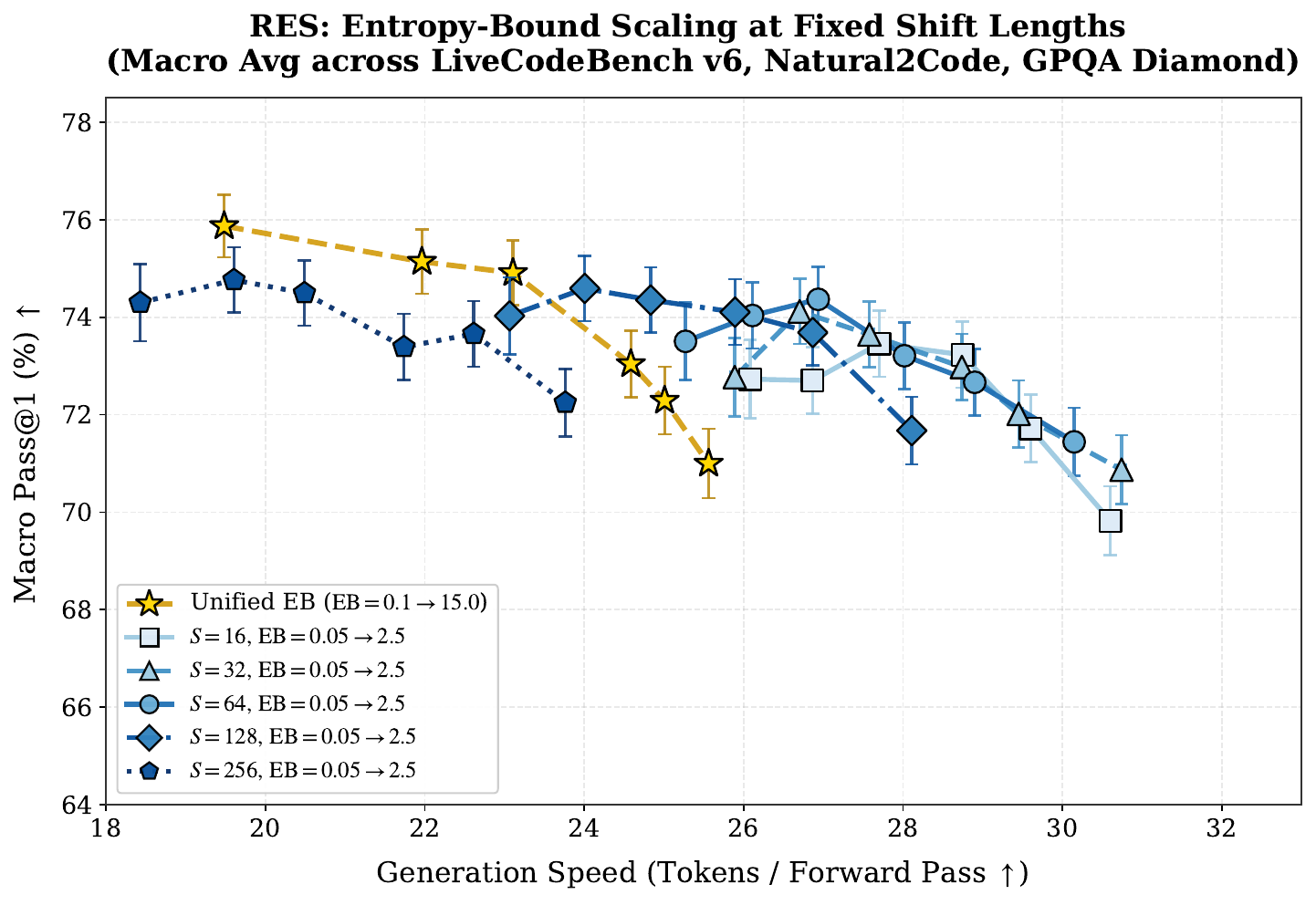}
    \caption{Frontiers obtained by varying both the minimum shift $\ell_{\min}$ in $F_{commit}$ (``S'' in the legend) and the entropy bound in $F_{route}$ for RES. Small shifts extend the frontier to higher throughput than Unified EB reaches.}
    \label{fig:dual-frontier}
\end{figure}

\paragraph{Entropy-Bound Sweep}
We sweep the entropy bound for RES with minimum shift $\ell_{\min}=32$ and for Unified EB, and show the frontiers in Figure~\ref{fig:minshift-16}.
Relaxing the entropy bound alone takes Unified EB to about 25.5 TPF, where macro Pass@1 has fallen to about 71\%.
At a similar speed RES holds about 74\%, and it reaches the same 71\% only at about 31 TPF.
The gain is in the high-throughput regime: below about 23 TPF Unified EB remains the better choice, so RES extends the frontier rather than dominating it.

\paragraph{Minimum Shift Sweep}
While Unified EB has one axis to trade off performance and speed, RES has the minimum shift in addition to the entropy bound.
Figure~\ref{fig:dual-frontier} shows the frontier for each minimum shift.
Minimum shifts of 16 to 64 give similar frontiers, between about 26 and 31 TPF.
Larger shifts move the frontier toward lower throughput: $\ell_{\min}=128$ covers about 23 to 28 TPF, and $\ell_{\min}=256$ about 18 to 24 TPF.
This direction is expected: a larger minimum shift requires a longer accepted prefix before commitment, moving the decision toward the block level and retaining positions in the denoising window for more computation.
At $\ell_{\min} = 256$, RES differs from Unified EB only in its position-based temperature schedule, so the remaining accuracy gap at low throughput comes from that schedule.

\subsection{Multi-Sample Performance}
RES also scales better with the number of samples (Figure~\ref{fig:macro_pass_at_k}).
Unified EB has the higher Pass@1, but RES with $\ell_{\min}=32$ overtakes it from $k \ge 6$ samples and maintains its advantage through $k=20$.
We hypothesize that the EB sampler avoids making ``difficult'' decisions, as it always accepts the lowest entropy positions first.
In contrast, the rolling mechanism along with the position-based temperature schedule enforces a left-to-right generation bias, which was shown in \citet{ni2026flexibility} to improve diversity across generations.

\section{Related Works}
\label{sec:adjacentworks}
This paper has three parts: a study of when a sampler should commit tokens, evidence that revising accepted tokens matters in uniform diffusion, and a rolling sampler that needs no training. Each part has close prior or concurrent work.

\subsection{Rolling Windows for Language}
Several methods move the denoising window forward as tokens finish.
DSB \citep{luo2026dsb} does this for masked diffusion.
Set Block Decoding \citep{gat2025set} fine-tunes an autoregressive model to decode masked tokens ahead with the entropy bound of \citet{ben2026accelerated}.
Concurrent with our work, Clock Diffusion \citep{schiff2026clock} trains a Gaussian diffusion model whose window rolls left to right and commits finished positions to the KV cache.
Its noise schedule depends on position, and our position-based temperature schedule plays the same role at inference time.
\citet{wu2023ar} paired diffusion with the left-to-right bias of language earlier still.
RES uses the same rolling mechanism, and we do not claim the mechanism as new.

\subsection{Freezing versus Revising}
In masked diffusion, an unmasked token stays fixed.
A rolling masked sampler such as DSB therefore never revises a token once it accepts it.
\citet{kim2025fine} fine-tune masked diffusion models so that they can revise tokens they have already revealed.

\subsection{What This Paper Adds}
Freezing accepted tokens in DiffusionGemma lowers macro accuracy from 73\% to 20\% at loose entropy bounds (Section~\ref{sec:too-early}).
A rolling sampler must therefore balance two goals: moving tokens into the KV cache early, and keeping them open to revision.
From the order in which tokens are accepted (Section~\ref{sec:too-late}), we design a commit rule with two properties.
Its minimum shift moves the sampler between block-level commitment and fully incremental commitment, which commits whenever it can.
Its entropy bound still trades accuracy for speed, as in the baseline.
Unlike Clock Diffusion, RES needs no training and runs on an existing model.

\subsection{Rolling Diffusion Beyond Text}
Beyond text generation, the idea of rolling diffusion has been studied extensively in other domains.
Within video generation, rolling diffusion was a natural mechanism to address the temporal nature of video generation and avoid performing diffusion on every frame simultaneously \citep{ruhe2024rolling}.
Many other works include similar mechanisms \citep{chen2024diffusion, song2025history, xie2025progressive}, and \citet{papa2026asyncpatch} carry the idea over to images by assigning different patches different levels of corruption.

\section{Conclusion}
We studied when a diffusion language sampler should commit a position to the KV cache.
In DiffusionGemma, freezing tokens upon acceptance degrades performance at higher entropy bounds, while block-level commitment spends much of each window on positions that have already converged.
RES commits a sufficiently long accepted prefix and keeps positions farther ahead open to revision.
It needs no training, and across seven benchmarks it increases tokens per forward pass from 19.3 to 25.5, while macro accuracy goes from 80.7\% to 78.8\%.
The block sampler remains better at low throughput.

\paragraph{Limitations}
Tokens per forward pass counts forward passes, not wall-clock time, although each RES pass processes the same number of positions as the baseline.
All experiments use one model, DiffusionGemma.

\paragraph{Future Work}
The largest cost of RES is on instruction following, where IFEval accuracy falls by 10.7\% (absolute).
One possible cause is that RES creates a left-to-right noise pattern that differs from the noise seen in training.
Training with a rolling noise pattern, as in Clock Diffusion \citep{schiff2026clock}, is a natural way to test and remove this cost.

\section*{Acknowledgments}
We thank Kelly He for thoughtful advice and input; Justin Deschenaux, Valentin De Bortoli, Alexandre Galashov, and Yazid Janati for all hands support; and Abel Gurung for insightful technical discussions.
\clearpage
\begingroup
\small
\setlength{\bibsep}{3pt}
\setlength{\parskip}{0pt}
\bibliography{main}
\bibliographystyle{main}
\endgroup

\clearpage
\appendix
\section{Appendix}

\subsection{Computation of Order Statistics}
\label{app:order-stats}
Here we provide details on how the order statistics in Figure~\ref{fig:wasted} (top row) are calculated.

\paragraph{Acceptance Event}
To track the ordering for the Unified EB sampler, we look at the first
acceptance per position, as the standard entropy bound of $0.1$ results in
positions rarely changing after the first acceptance. Concretely, for a
canvas of $b$ positions $X_1,\dots,X_b$ denoised over $T$ steps, we record
for each position the step at which it is first accepted,
\begin{equation}
  c_i \;=\; \min\bigl\{\, t \in \{1,\dots,T\} \;:\; X_i \text{ is accepted at step } t \,\bigr\},
\end{equation}
setting $c_i = T$ for the rare positions never accepted. Each generated
canvas therefore yields an acceptance-step vector $c \in \{1,\dots,T\}^{b}$. A
purely autoregressive sampler would produce $c_i = i$, so we measure how
closely $c$ tracks the left-to-right index. These are first-acceptance times, not the times at which positions enter the KV cache.

\paragraph{Resolving Ties}
Since our goal is to compare the ranking of positions based on acceptance
step, there must be a way to resolve the ties between tokens that are
accepted on the same step. To resolve this, we take the approach of
averaging the order ranking across slots that are tied, assigning each
position the fractional midrank
\begin{equation}
  r_i \;=\; \bigl|\{\,k : c_k < c_i\,\}\bigr|
        \;+\; \tfrac{1}{2}\bigl(1 + \bigl|\{\,k : c_k = c_i\,\}\bigr|\bigr).
\end{equation}
Given the list of acceptance steps $c = [1, 3, 1, 2, 3, 3]$, we first sort them
in ascending order, $[1, 1, 2, 3, 3, 3]$. The positions accepted on the
first step occupy the first and second ranking slots, so both are assigned
the average rank $1.5$; likewise the three positions accepted on the third
step occupy slots four through six and are each assigned $5$. Returning
these ranks to their original positions yields
$r = [1.5,\, 5,\, 1.5,\, 3,\, 5,\, 5]$. Averaging within tie groups keeps
tied positions exchangeable, so the statistic does not depend on an
arbitrary tie-breaking order, and preserves the rank sum
$\sum_i r_i = b(b+1)/2$.

\paragraph{Spearman $\rho$}
Given the vector of rankings, we compare it against the standard
left-to-right rankings $[1, 2, \dots, b-1, b]$ and take the Pearson
correlation coefficient to be the Spearman rank correlation $\rho$. Writing
$\pi_i = i$ for the reference ranking,
\begin{equation}
  \rho \;=\;
  \frac{\sum_{i=1}^{b} (\pi_i - \bar{\pi})(r_i - \bar{r})}
       {\sqrt{\sum_{i=1}^{b} (\pi_i - \bar{\pi})^{2}}\;
        \sqrt{\sum_{i=1}^{b} (r_i - \bar{r})^{2}}},
  \qquad \bar{\pi} = \bar{r} = \frac{b+1}{2}.
\end{equation}
We use this general form rather than the familiar
$1 - 6\sum_i d_i^{2}/\bigl(b(b^{2}-1)\bigr)$, which is valid only when the
ranks are untied.

\paragraph{Kendall $\tau$}
Spearman rank correlation may not reflect the true behavior, as it is
quadratic in terms of the differences per position. Thus we also calculate
the Kendall $\tau$, as it counts inversions and is thus less sensitive to
large differences. More specifically, for every position pair $(X_i, X_j)$
with $i < j$, the pair is \emph{concordant} if $X_i$ was accepted earlier
than $X_j$, \emph{discordant} if $X_j$ was accepted earlier, and
\emph{tied} if both were accepted on the same step. Over all
$n_0 = \binom{b}{2}$ pairs we count
\begin{equation}
  n_c = \bigl|\{(i,j) : i<j,\; c_i < c_j\}\bigr|, \quad
  n_d = \bigl|\{(i,j) : i<j,\; c_i > c_j\}\bigr|, \quad
  n_t = n_0 - n_c - n_d .
\end{equation}
Because the reference ranking is strictly increasing, it contributes no
ties of its own, and every tied pair is tied in acceptance step. We report the
tie-corrected $\tau_b$, which normalises by the pairs whose order is
resolvable at all,
\begin{equation}
  \tau_b \;=\; \frac{n_c - n_d}{\sqrt{n_0\,(n_0 - n_2)}},
  \qquad
  n_2 \;=\; \sum_{s=1}^{T} \binom{m_s}{2},
\end{equation}
where $m_s$ is the number of positions first accepted at step $s$. We
additionally report the pairwise left-to-right agreement rate, which counts
ties as half-agreements,
\begin{equation}
  P_{\mathrm{l2r}} \;=\; \frac{n_c + \tfrac{1}{2} n_t}{n_0},
\end{equation}
equal to $50\%$ under a random acceptance order and $100\%$ under a purely
left-to-right one.

\paragraph{KDE Estimate}
To get the shape of the distribution, we use the default KDE estimator in matplotlib.

\end{document}